\documentclass{article}

\PassOptionsToPackage{numbers, compress}{natbib}

\usepackage[main,final]{neurips_2026}
\usepackage{graphicx} 
\usepackage[utf8]{inputenc} 
\usepackage[T1]{fontenc}    
\usepackage{hyperref}       
\usepackage{url}            
\usepackage{booktabs}       
\usepackage{amsfonts}       
\usepackage{nicefrac}       
\usepackage{microtype}      
\usepackage{xcolor}         
\usepackage{enumitem}       
\usepackage{amsmath}        
\usepackage{subcaption}     
\usepackage{float}          
\usepackage{wrapfig}        
\usepackage{multirow}

\usepackage[ruled,vlined]{algorithm2e}    
\usepackage{caption}

\title{Escaping Path Mirages in Offline Goal-Conditioned Reinforcement Learning}
\author{%
  Junhyeon Bae\textsuperscript{*} \quad
  Jaebak Hwang\textsuperscript{*} \quad
  Gwanwoo Choi \quad
  Minung Kim \quad
  Seungyul Han\textsuperscript{\dag} \\
  Graduate School of Artificial Intelligence, UNIST \\ 
  Ulsan, South Korea \\
  \texttt{\{junhyeonbae,milk,cgw1999,minungkim,syhan\}@unist.ac.kr}
}
\date{March 2026}

\begin{document}

\maketitle

{\let\thefootnote\relax
\footnotetext{\textsuperscript{*}Equal contribution. \quad \textsuperscript{\dag}Corresponding author: Seungyul Han.}}

\begin{abstract}
Offline goal-conditioned reinforcement learning remains challenging in stochastic long-horizon settings, where compounding value estimation errors hinder reliable goal reaching. While prior methods have sought to address this challenge through various approaches, a key challenge remains path mirage, where agents overcommit to spuriously successful trajectories in offline datasets induced by stochasticity, often leading to failure on difficult tasks. To address this issue, we propose Branch-aware Graph Planning (BGP), which captures stochastic branching structures by identifying high-variability points as graph nodes and designing goal-conditioned edges with segmented learning to avoid unreliable branches and reduce unnecessary stochasticity along paths. As a result, BGP enables reliable goal reaching even in highly stochastic long-horizon environments, and experiments on diverse OGBench tasks show that it substantially outperforms prior state-of-the-art offline goal-conditioned RL methods. Our implementation is available at \url{https://umlic.github.io/papers/BGP/}.
\end{abstract}

\section{Introduction}
\label{sec:intro}

Offline reinforcement learning (RL) has emerged as a practical paradigm for learning from static datasets when additional environment interaction is costly or unsafe~\citep{kostrikov2021offline,kumar2020conservative,yeom2024exclusively}. Beyond reproducing behaviors from data, many real-world applications require agents to reach diverse user-specified goals, making offline goal-conditioned reinforcement learning (GCRL) a compelling extension of offline RL~\citep{andrychowicz2017hindsight,park2023hiql}. Despite recent progress in hierarchical subgoal learning~\citep{ajay2020opal,li2022hierarchical}, graph-based planning~\citep{baek2025graph}, and temporal distance or reachability learning~\citep{eysenbach2022contrastive,wang2023optimal,myers2025offline}, offline GCRL remains fundamentally challenging in stochastic long-horizon domains, where compounding value estimation errors can undermine reliable goal reaching.

A key challenge arises from stochastic branching structures implicit in offline datasets. In stochastic environments, the same local decision can lead to multiple divergent futures, while rare lucky outcomes can create deceptively successful trajectories that appear reliably goal-reaching even when they are not. We refer to this failure mode as \textit{path mirage}, where agents overcommit to such misleading trajectories and propagate overly optimistic value estimates through unreliable branches. This issue becomes particularly severe in long-horizon settings, where local branching errors accumulate over time and derail planning far from the true goal. Although existing methods improve trajectory composition or exploit connectivity in offline data for long-horizon reasoning~\citep{park2023hiql,baek2025graph,wang2023optimal,myers2025offline}, they often treat that connectivity as uniformly trustworthy, leaving stochastic branching ambiguity largely unresolved.

To address this challenge, we propose \textbf{Branch-aware Graph Planning (BGP)}, a framework that explicitly reasons about stochastic branching during planning and value learning by identifying high-variability branching points as graph landmarks and constructing goal-conditioned edges that encode reliable progress toward the goal, enabling agents to avoid deceptive paths induced by path mirage. In addition, BGP incorporates segmented learning that limits value propagation across unreliable stochastic branches to reduce stochasticity-induced errors and mitigate long-horizon error accumulation. Together, these components reformulate offline GCRL as branch-aware graph planning over unreliable connectivity, enabling agents to reach distant goals even in highly stochastic environments. Through extensive experiments on challenging OGBench~\citep{park2024ogbench} tasks, we show that BGP consistently outperforms prior state-of-the-art offline GCRL methods, with particularly strong gains on stochastic long-horizon problems where path mirage poses a major challenge.

\section{Related Work}
\label{sec:related works}

\paragraph{Offline goal-conditioned reinforcement learning.}
\label{subsec:related_offline_gcrl}

Offline goal-conditioned reinforcement learning has developed several complementary directions to address extrapolation error and long-horizon generalization in static datasets. One line focuses on value learning, using conservative regularization~\citep{chebotar2021actionable,ke2025conservative} or expectile regression~\citep{park2023hiql,ahn2025option} to mitigate overestimation. Another replaces temporal-difference learning with supervised formulations based on behavior cloning~\citep{yang2022rethinking,emmons2021rvs}, sequence modeling~\citep{janner2021offline,chen2021decision}, or goal-conditioned trajectory generation~\citep{ajay2022conditional,kim2024stitching,rouxel2025extremum,koirala2025flow,chae2026flow}. Other approaches model reachability through contrastive objectives~\citep{eysenbach2022contrastive,myers2024learning}, f-divergence optimization~\citep{ma2022offline,sikchi2023score}, or quasi-metric learning~\citep{wang2023optimal,myers2025offline}. More recently, hierarchical methods further decompose long-horizon tasks into latent subgoals or skills to improve compositional generalization~\citep{park2023hiql,ajay2020opal,li2022hierarchical}.

\vspace{-0.1in}
\paragraph{Graph-based planning for goal-conditioned RL.}
\label{subsec:related_graph}

Graph-based methods structure long-horizon decision making through explicit connectivity and search. In online RL, some approaches directly perform graph search over learned connectivity structures for waypoint planning~\citep{eysenbach2019search,huang2019mapping,zhang2021world}, while others construct topological or latent graphs to support subgoal planning and long-horizon reasoning~\citep{savinov2018semi,shah2021ving,kim2021landmark,lee2022dhrl,hwang2025strict}. In offline settings, graph-based approaches exploit data-supported connectivity to construct planning structures for long-horizon goal reaching~\citep{baek2025graph,kobanda2026adaptive,liangbottleneck}. Similar in spirit, our method also builds on graph-based planning, but extends it beyond connectivity alone by explicitly reasoning about reliability under stochastic branching, identifying stochastic landmarks as core nodes and constructing branch-aware goal-conditioned edges for reliable long-horizon planning.

\vspace{-0.1in}
\paragraph{Uncertainty-aware reinforcement learning.}
\label{subsec:related_uncertainty}

A large body of work addresses stochasticity through uncertainty-aware or risk-sensitive learning. In online RL, stochasticity is modeled through distributional value estimation~\citep{bellemare2017distributional,dabney2018distributional,dabney2018implicit}, uncertainty-aware model learning~\citep{chua2018deep,kurutach2018model,janner2019trust}, or robust planning under predictive uncertainty. In offline RL, stochastic risks are often handled through pessimistic penalties~\citep{kumar2020conservative,bai2022pessimistic,cheng2022adversarially}, uncertainty-aware model constraints~\citep{yu2020mopo,kidambi2020morel,wu2021uncertainty}, or risk-sensitive return estimation~\citep{ma2021conservative,urpi2021risk}. While these approaches primarily use uncertainty to regularize value learning or control risk, our method instead uses stochasticity as a structural signal for identifying branching landmarks and guiding reliable graph construction.

\section{Preliminaries}
\label{sec:preliminary}

\subsection{Offline Goal-Conditioned RL Setup}
\label{subsec:preliminary_setup}

We consider a goal-conditioned Markov decision process (GCMDP) defined by $(\mathcal S,\mathcal G,\mathcal A,\mathcal P,r,\gamma)$, where $\mathcal S$ denotes the state space, $\mathcal G$ the goal space, $\mathcal A$ the action space, $\mathcal P$ the transition dynamics, $r$ the goal-conditioned reward function, and $\gamma\in[0,1)$ the discount factor. In this work, we adopt the standard setting $\mathcal G=\mathcal S$, where goals are specified directly in the state space. In the offline setting, the agent learns a policy $\pi(a_t \mid s_t, g)$ to maximize the expected return $\mathbb{E}[\sum_t \gamma^t r(s_t, g)]$ solely from a dataset $\mathcal D$ of trajectories without additional environment interaction. We consider a goal-reaching reward defined as $r(s_t,g)=0$ if $s_t=g$ and $r(s_t,g)=-1$ otherwise.

Long-horizon GCMDPs in offline settings pose challenges for value estimation, motivating value-based methods such as Implicit Q-Learning (IQL)~\citep{kostrikov2021offline}, which employs expectile regression for stability. In goal-conditioned settings, the value function $V_\theta(s,g)$ can be learned as
\begin{equation}
\mathcal L_V(\theta) 
= 
\mathbb E_{(s,s')\sim\mathcal D} 
\Big[ 
\ell_\kappa^2 
\big( 
r(s,g)+\gamma V_{\bar\theta}(s',g)-V_\theta(s,g) 
\big) 
\Big],
\label{eq:hiql_value}
\end{equation}
where $\ell_\kappa^2(u)=|\kappa-\mathbb I(u<0)|u^2$ denotes the expectile loss (state-value formulation). During training, the goal $g$ is sampled from a mixture of the current state, a future state from the same trajectory, and a random state from the full offline dataset. Under this reward, the learned value can be interpreted as the negative temporal distance to the goal. Representative methods include GCIQL, which learns a goal-conditioned policy by maximizing this value, and HIQL~\citep{park2023hiql}, which extends it hierarchically. We use these methods as baselines, and detailed training objectives are provided in Appendix~\ref{subsecapp:other_baseline}.

\subsection{Graph-based Planning for Goal Conditioned MDPs}
\label{subsec:preliminary Graph-based RL}
Graph-based planning abstracts the goal space into a directed graph $G=(\mathcal V,\mathcal E)$ consisting of a landmark set $\mathcal V$ and edge set $\mathcal E$. Each landmark node corresponds to a selected state, so we regard $\mathcal V$ as a subset of $\mathcal S$. Landmark selection can be performed in a representation space using a mapping $\phi:\mathcal S\rightarrow\mathcal Z$, such as the identity mapping, a PCA projection~\citep{abdi2010principal}, or a learned Hilbert representation~\citep{park2024foundation}. Landmark nodes are typically selected to provide broad coverage of the goal space, for example via farthest-point sampling~\citep{arthur_kmeans_2007} or other task-specific criteria. For each connected edge $(v,w)\in\mathcal E$, where $v,w\in\mathcal V$, an edge cost $e(v,w)$ quantifies the reachability from node $v$ to node $w$, and is commonly defined using distance-based metrics or learned goal-conditioned value functions. Given the constructed graph, planning toward a target goal $g$ can be performed via shortest-path search, e.g., Dijkstra's algorithm~\citep{dijkstra1959note}, yielding a waypoint sequence $\mathcal{W}=(w_0, \ldots, w_K)$ where $w_0 = s$ and $w_K = g$. Each waypoint $w_i\in \mathcal V$ is sequentially used as an intermediate goal for the
underlying goal-conditioned policy, advancing to $w_{i+1}$ once the current waypoint is
reached, e.g., when $\|\phi(s_t)-\phi(w_i)\|<\epsilon$ for a threshold $\epsilon$, where
$\phi$ is the state representation introduced above. We extend this graph-based planning framework with branch-aware landmark construction and goal-conditioned edges.

\section{Methodology}
\label{sec:method}

\subsection{Motivation: Path Mirage in Offline Goal-Conditioned RL}
\label{subsec:motivation}

\begin{figure}[H] 
    \centering
    \includegraphics[width=1.0\linewidth]{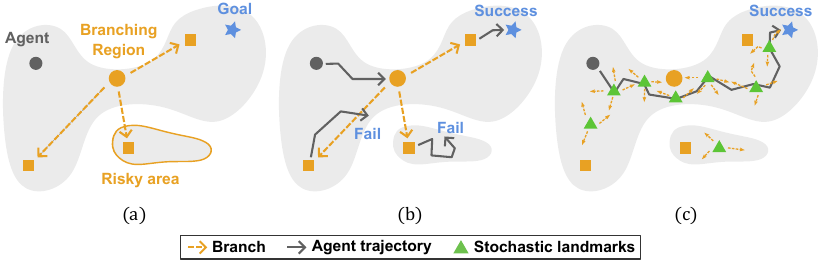}
    \caption{
    Path mirage in offline GCRL. 
    (a) Offline trajectories can pass through branching regions where future outcomes diverge toward success, failure, or risky areas. 
    (b) Prior methods can be misled by deceptively successful paths, resulting in path mirages with occasional successes but persistent failures. 
    (c) Our branch-aware graph planning identifies branching points near branching regions as landmarks and selects reliable goal-conditioned routes that avoid risky branches.
    }
    
    \vspace{-1em}
    \label{fig:motivation}
\end{figure}

Long-horizon goal-conditioned tasks often admit multiple branching paths toward a goal, where trajectories originating from the same or nearby states can lead to drastically different future outcomes. Fig.~\ref{fig:motivation}(a) illustrates an extreme yet representative case, where trajectories in the offline dataset pass through branching regions whose outgoing paths may lead either near the goal, far away from it, or into irreversible dead zones. A central challenge is that offline datasets typically contain only unlabeled trajectories, without indicating which goal a trajectory was intended to reach or whether it reflects reliable progress or latent failures. As a result, such branching structure is not explicitly recognized. Although prior offline GCRL methods may partially learn to avoid low-value regions, value estimation alone is often insufficient in long-horizon settings, leading to a \emph{path mirage} phenomenon: as shown in Fig.~\ref{fig:motivation}(b), the agent is misled into treating stochastic branching points as uniquely promising paths to the goal, while failing to account for unreliable or risky outcomes. Consequently, policies repeatedly route through such deceptive states, achieving occasional successes while remaining prone to persistent failures.

To address this, we propose \emph{Branch-aware Graph Planning}, which explicitly treats highly stochastic branching points as landmarks and reasons about whether paths emanating from them are reliable for a given goal. Rather than planning over connectivity alone, our method uses a goal-dependent edge design to assess whether paths branching from a landmark facilitate or hinder reaching the target goal, thereby steering planning away from risky branches. As illustrated in Fig.~\ref{fig:motivation}(c), this yields a graph whose landmarks capture future branching structure and whose edges favor reliable goal-directed routes, allowing the agent to navigate stochastic long-horizon environments more cautiously and reach distant goals more reliably. We detail this branch-aware planning framework in the following sections.

\subsection{Branch-aware Graph Construction}
\label{sec:graph}

We construct a branch-aware goal-conditioned graph that enables reliable planning under stochastic branching via stochastic landmark identification and goal-conditioned edge design.

\vspace{-.5em}

\paragraph{Stochastic landmark identification.} 

To capture branching structures in the dataset, we identify states from which future trajectories can diverge under stochastic transitions, as these determine the reliability of long-horizon planning. We quantify this by assigning each state $s_t \in \mathcal{D}$ a multi-step predictive uncertainty score $\mathcal{U}^{(n)}$ defined as the discounted sum of predictive entropies over $n$ steps:
\begin{equation}
\mathcal{U}^{(n)}(s_t)
= \sum_{l=0}^{n-1} \gamma^l \,
H\!\left(p_\psi(y_{t+l} \mid s_t, y_{t:t+l-1})\right),
\label{eq:uncertainty_score}
\end{equation}
where $H$ denotes entropy, $y_t := (a_t, s_{t+1})$ denotes the joint action-state transition, and $p_\psi$ is an autoregressive model conditioned on $s_t$ and previously predicted variables. In practice, $p_\psi$ is parameterized as a decoder-only Transformer with Gaussian outputs, enabling entropy computation. This formulation captures multi-step trajectory ambiguity rather than one-step noise, making $\mathcal{U}^{(n)}(s_t)$ a proxy for stochastic branching; in our experiments, we set $n=5$.

We then construct graph landmarks from the top-$\alpha$ percentile of high-uncertainty states, which capture branching regions in the dataset. We apply K-means++ clustering~\citep{arthur_kmeans_2007} to representations space $\mathcal Z$. We use a PCA projection for state-based observations and a Hilbert representation for visual observations. Let $\{\mathcal{C}_i^\alpha\}_{i=1}^N$ denote the resulting clusters with centers $\{c_i\}_{i=1}^N$. To ensure full coverage, we assign each state to its nearest cluster center in $\mathcal{Z}$:
\begin{equation}
\mathcal{C}_i=\mathcal{C}_i^\alpha \cup\left\{s \in \mathcal{D} \setminus \bigcup_{j=1}^N \mathcal{C}_j^\alpha  \;\middle|\; i=\operatorname*{arg\,min}_{k\in\{1,\ldots,N\}}
\|\phi(s)-\phi(c_k)\|_2\right\}.
\end{equation}

\begin{wrapfigure}[17]{r}{0.53\linewidth}
    \centering
    \vspace{-15pt}
    \includegraphics[width=.95\linewidth]{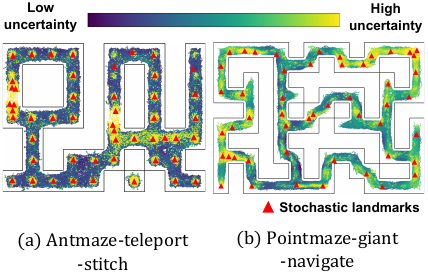}
    \caption{Visualization of uncertainty scores and the resulting landmarks on two OGBench tasks. High-uncertainty regions correspond to significant trajectory branching, leading to denser landmark selection.}
    \label{fig:uncertainty}
\end{wrapfigure}

Since cluster centers may not correspond to actual states, we define each landmark node $v_i$ as the state closest to the cluster center $c_i$ in the representation space $\mathcal Z$. This yields a set of landmarks $\mathcal{V} = \{v_1, \dots, v_N\}$ that capture the branching structure of the dataset. Details of the Transformer training are provided in Appendix~\ref{subsecapp:graph_construction}.

Fig.~\ref{fig:uncertainty} visualizes uncertainty scores and the resulting landmarks across OGBench tasks. Landmarks concentrate in regions with significant trajectory branching and remain sparse where branching is limited. Importantly, they are \emph{not required to be uniformly distributed}, as such sparse regions typically correspond to parts of the state space that do not induce large uncertainty in future trajectories.

\paragraph{Goal-conditioned edge construction.}

Unlike prior graph-based GCRL methods that rely on pairwise connectivity, we design edges to explicitly account for stochastic branching and goal-directed progress. Rather than evaluating a transition $(v_i, v_j)$ locally, we consider all possible downstream branches after visiting $v_j$, defined as
\[
\mathcal{R}(v_j)
:=
\left\{
\mathcal{C}_k \;\middle|\;
v_k \in \mathcal{V},\;
\exists\, s \in \mathcal{C}_j \text{ such that a trajectory from } s \text{ reaches } \mathcal{C}_k
\right\},
\]
which characterizes the branching outcomes induced by $v_j$. We then define the edge cost as
\begin{equation}
e_g(v_i, v_j)
=
\mathbb{E}_{s_i \sim \mathcal{C}_i,\; s_k \sim \mathcal{R}(v_j)}
\left[
\bigl(d(s_k, g)-d(s_i, g)\bigr)_+
\right],
\label{eq:edge}
\end{equation}
where $d(s, g)$ denotes a distance to the goal that reflects reachability, and $d=\infty$ if unreachable. Based on $e_g$, we also define the edge set as $\mathcal{E}_g
=
\left\{
(v_i, v_j) \;\middle|\;
e_g(v_i, v_j) < \infty
\right\}.$ 
This formulation evaluates whether transitioning from $v_i$ to $v_j$ leads, under all possible branches, to progress toward the goal. If the downstream trajectories reduce the distance to the goal, the edge cost becomes negative; otherwise, it incurs a positive penalty. Crucially, if any reachable branch leads to significantly worse outcomes or fails to reach the goal, the corresponding edge is assigned a large cost or effectively removed. 

\begin{figure}[t]
    \centering
    \includegraphics[width=0.9\linewidth]{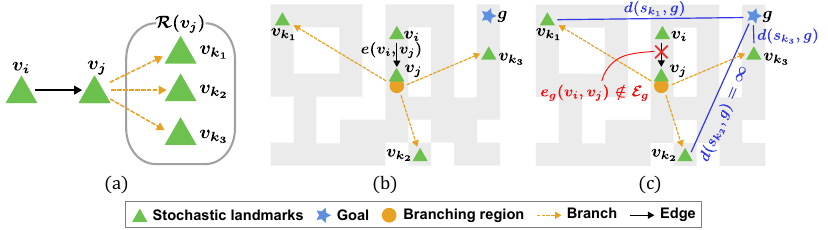}
\caption{
(a) Illustration of downstream branches $\mathcal{R}(v_j)$. 
(b)-(c) Edge comparison on a teleport task in OGBench, where $v_j$ (near the teleport) induces stochastic branching. 
(b) A prior edge $e(v_i, v_j)$ favors the transition due to local proximity, whereas 
(c) our goal-conditioned edge $e_g$ evaluates all branches $\{v_{k_1}, v_{k_2}, v_{k_3}\}\in \mathcal{R}(v_j)$ and avoids selecting $v_j$ since one branch ($v_{k_2}$) leads to a dead zone, i.e., $d(s_{k_2},g)=\infty$ for $s_{k_2}\in\mathcal{C}_{k_2}$.}
    \label{fig:figure4}
\end{figure}

Fig.~\ref{fig:figure4} illustrates the branching structure $\mathcal{R}(v_j)$ and the effectiveness of our edge design on a teleport task in OGBench, where $v_j$ corresponds to a teleport point such that transitioning to $v_j$ may stochastically lead to multiple outcomes, including dead zones from which the goal is unreachable. Fig.~\ref{fig:figure4}(b) shows that a goal-agnostic edge design $e(v_i, v_j)$ treats such transitions favorably due to local proximity, causing the planner to traverse these risky connections, whereas Fig.~\ref{fig:figure4}(c) demonstrates that our goal-conditioned edge $e_g$ evaluates all downstream branches of $v_j$ and penalizes transitions that may lead to unsafe or non-goal-reaching regions, enabling the planner to avoid risky transitions and select paths that reliably reach the goal under stochastic dynamics.

\subsection{Stability-oriented Distance Design and Graph Planning}
\label{sec:planning}

In the proposed branch-aware graph $\mathcal{G}_g$, we require a distance function that reflects reachability. A natural choice is the policy-induced discounted temporal distance $d^\pi(s,g):=-V^\pi(s,g)=\mathbb{E}_{\pi}\bigl[\textstyle\sum_{t=0}^{\tau_g-1}\gamma^t \mid s_0=s,g\bigr]$, where $\tau_g$ is the first hitting time of goal $g$ under $\pi$. Under the per-step reward of $-1$ before reaching $g$ and $0$ upon reaching it, this quantity is the expected discounted number of steps to the goal. In practice, we estimate it using $-V_\theta(s,g)$, with $V_\theta$ learned via Eq.~(\ref{eq:hiql_value}). However, this becomes unreliable in long-horizon stochastic settings, where it must extrapolate over distant and highly uncertain trajectories. Instead, BGP adopts a \textit{two-phase graph planning}, consisting of two graphs that share the landmark set $\mathcal{V}$ but have distinct edge sets. The \textbf{pre-graph} $\mathcal{G}_p$ provides a stability-oriented distance between landmarks, while the \textbf{branch-aware graph} $\mathcal{G}_g$ uses this distance in its edge costs and serves as the graph for actual waypoint planning. We construct a pre-graph $\mathcal{G}_p = (\mathcal{V}, \mathcal{E}_p)$ with edge weights $e_p(v, w) = -V_\theta(v, w)$, which capture locally reliable transitions between neighboring landmarks. We then define the distance by composing these local predictions along a path:
\[
d(s, g) = \sum_{i=0}^{K-1} e_p(w_i, w_{i+1}) = - \sum_{i=0}^{K-1} V_\theta(w_i, w_{i+1}),
\]
where $(w_0, \ldots, w_K)$ is the shortest path from $w_0 = s$ to $w_K = g$ on $\mathcal{G}_p$, computed via Dijkstra’s algorithm, and $d(s, g) = \infty$ if no such path exists.  
This formulation replaces a single long-horizon value estimate with a composition of short-horizon, locally accurate predictions, enabling more reliable reachability estimation under stochastic dynamics.

Based on the goal-conditioned graph $\mathcal{G}_g$ with distance $d$, we perform \textbf{Branch-aware Graph Planning (BGP)} by computing the shortest waypoint path $(\tilde{w}_0, \ldots, \tilde{w}_{K'})$ using Dijkstra's algorithm with edge costs $e_g$. The agent is then guided to sequentially reach these waypoints as intermediate goals, enabling reliable goal-reaching under stochastic branching. To learn policies for reaching each waypoint, we can in principle adopt any value-based offline GCRL method. In practice, we use the representative IQL-based approach HIQL~\cite{park2023hiql} as our baseline, as it is designed to handle long-horizon execution.

\subsection{Segmented Learning}
\label{sec:segmented Learning}

Unlike prior methods that learn $V_\theta(s, g)$ over full trajectories, BGP only requires reaching nearby landmarks. To mitigate compounding value-estimation errors under stochastic transitions, we employ segmented learning to train the value function and policy on shorter trajectory segments defined by cluster transitions. Specifically, we partition each $k$-th offline trajectory $\tau^{(k)}$ in $\mathcal{D}$ into trajectory segments $\tau_j^{(k)}$ at boundaries where the cluster assignment changes. We then concatenate $M$ consecutive segments to form a local training window, $\tilde{\tau}_j^{(k)} := \left(\tau_j^{(k)}, \tau_{j+1}^{(k)}, \ldots, \tau_{j+M-1}^{(k)}\right)$. During offline learning, we sample both transitions and their target goals exclusively from the same window, i.e., $(s_t, a_t, s_{t+1}) \sim \tilde{\tau}_j^{(k)}$ and $g \sim \tilde{\tau}_j^{(k)}$, thereby limiting the effective horizon of value propagation and reducing the accumulation of value-estimation errors across stochastic boundaries. Thus, $M$ controls the trade-off between local trajectory coverage and learning stability. 

\begin{figure}[!t]
    \centering
    \begin{minipage}[t]{0.48\textwidth}
        \vspace{0pt}
        \includegraphics[width=\linewidth]{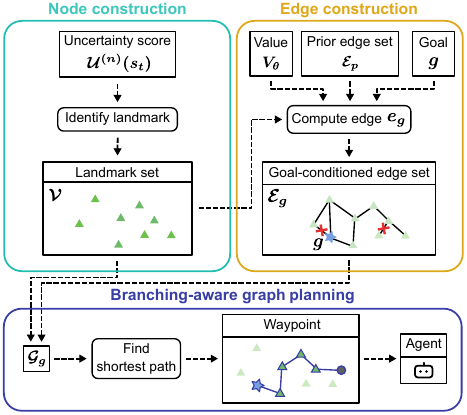}
        \vspace{-0.2in}
        \caption{The proposed BGP framework}
        \label{fig:sasl}
    \end{minipage}
    \hfill
    \begin{minipage}[t]{0.51\textwidth}
        \vspace{0pt}
        \begin{algorithm}[H]
\small
\caption{Branch-aware Graph Planning}
\label{alg:bgp}
\DontPrintSemicolon
\SetAlgoLined

\textbf{Require:} Offline dataset $\mathcal{D}$, goal $g$\;

\BlankLine
\textbf{(1) Landmark node construction}\;
Compute uncertainty score $\mathcal{U}^{(n)}(s)$ using $p_\psi$\;
Construct landmark set $\mathcal{V}=\{v_i\}_{i=1}^N$\;

\BlankLine
\textbf{(2) Goal-conditioned edge construction}\;
Train policy $\pi$ and value function $V_\theta$ via offline GCRL, and construct pre-graph $\mathcal{G}_p$\;
Given $g$, compute goal-conditioned edge costs $e_g(v_i, v_j)$ for all $v_i, v_j \in \mathcal{V}$\;
Construct graph $\mathcal{G}_g=(\mathcal{V}, \mathcal{E}_g)$\;

\BlankLine
\textbf{(3) Offline RL with graph planning}\;
Compute shortest waypoint path $(\tilde{w}_0, \ldots, \tilde{w}_{K'})$ on $\mathcal{G}_g$ via Dijkstra's algorithm\;
Execute $\pi$ to reach $g$ via waypoints\;

\end{algorithm}
    \end{minipage}
\end{figure}

Overall, BGP explicitly accounts for stochastic branching, avoids risky transitions, and enables reliable goal-reaching. Fig.~\ref{fig:sasl} illustrates the proposed framework, while Algorithm~\ref{alg:bgp} summarizes the complete procedure. Further implementation details on graph construction and segmented learning are provided in Appendix~\ref{secapp:implementation_details}.

\label{subsec:experiment_setup}
\begin{figure}[t]
    \vspace{-1em}
    \centering
    \includegraphics[width=1.0\linewidth]{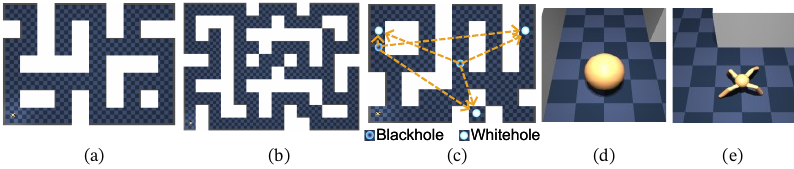}
   \caption{Visualization of OGBench tasks. (a)--(c) show three maze layouts: \texttt{large}, \texttt{giant}, and \texttt{teleport}. In the \texttt{teleport} layout, two blackholes teleport the agent to one of three whiteholes uniformly at random. (d) and (e) show the \texttt{pointmaze} and \texttt{antmaze} tasks, respectively.}
    \vspace{-1em}
    \label{fig:env}
\end{figure}

\section{Experiment}

\subsection{Experimental Setup}
\label{sec:experimental setup}
We evaluate BGP on OGBench~\citep{park2024ogbench}, a widely used offline GCRL benchmark, as shown in Fig.~\ref{fig:env}. We consider three challenging maze layouts: \texttt{large}, \texttt{giant}, and \texttt{teleport}. The \texttt{large} and \texttt{giant} layouts induce branching paths with substantial stochasticity, while \texttt{teleport} adds blackholes that randomly teleport the agent to one of several locations. Our main experiments cover \texttt{antmaze} for quadruped control, \texttt{pointmaze} for point-mass navigation, and \texttt{visual-antmaze} for pixel-based quadruped control. Appendix~\ref{subsecapp:additional_tasks} presents additional performance on higher-dimensional humanoid control and manipulation tasks. OGBench provides three dataset types: \texttt{navigate} contains direct goal-reaching trajectories, \texttt{stitch} requires composing disconnected fragments into long-horizon paths, and \texttt{explore} offers a highly noisy data distribution, each with five target goals per layout. More detailed descriptions of the environments and goal specifications are provided in Appendix \ref{subsecapp:environment_detail}.

We compare our method with a broad set of prior offline GCRL baselines. As primary baselines, we consider IQL-based GCRL baselines, \textbf{GCIQL} and \textbf{HIQL}~\citep{park2023hiql}. We further include \textbf{QRL}~\citep{wang2023optimal} and \textbf{TMD}~\citep{myers2025offline}, which learn quasimetric goal-reaching or temporal distances; \textbf{CGCIVL}~\citep{ke2025conservative}, which introduces a conservative value-learning objective to mitigate overestimation in cross-trajectory stitching; and \textbf{CRL}~\citep{eysenbach2022contrastive}, which uses contrastive representation learning. We also include \textbf{GAS}~\citep{baek2025graph}, which formulates subgoal selection as a graph-based planning approach. For each method, we report the official results on tasks where they are available; for the remaining cases, we reproduce results using the authors' code with the hyperparameter tuning protocols specified in their papers. For the proposed BGP, we report the best configuration over the searched key hyperparameters. For fair comparison, all methods are trained for 500K steps on pixel-based tasks and 1M steps otherwise. Detailed descriptions of the baselines and our hyperparameter settings are provided in Appendix~\ref{subsecapp:other_baseline}.

\label{sec:ogbench_performance}
\begin{table*}[!t]
    \centering
    \scriptsize
    \setlength{\tabcolsep}{2pt}
    \renewcommand{\arraystretch}{1.2}
    \newcommand{\best}[1]{\textcolor[rgb]{0.00,0.58,0.62}{\textbf{#1}}}
    \newcommand{\std}[1]{\textsuperscript{($\pm$#1)}}
    \caption{Performance comparison on OGBench. We report the average success rate (\%) with standard deviation ($\pm$) over 5 seeds. The best results in each row are highlighted in \best{bold}.}
    \label{tab:ogbench_performance}
    \begin{tabular*}{\textwidth}{@{\extracolsep{\fill}}llcccccccc@{}}
        \toprule
        \textbf{Env} & \textbf{Dataset} & \textbf{HIQL} & \textbf{GCIQL} & \textbf{QRL} & \textbf{CRL} & \textbf{TMD} & \textbf{CGCIVL} & \textbf{GAS} & \textbf{BGP(Ours)} \\
        \midrule
        \multirow{8}{*}{\texttt{antmaze}}
        & \texttt{large-navigate}   & 91\std{2}  & 34\std{4}  & 75\std{6}   & 83\std{4}   & 82\std{2}  & 91\std{2}   & \best{93\std{1}}  & 90\std{1} \\
        & \texttt{large-stitch}     & 67\std{5}  & 7\std{2}   & 18\std{2}   & 11\std{2}   & 37\std{3}  & 79\std{3}   & \best{96\std{1}}  & 93\std{1} \\
        & \texttt{large-explore}    & 4\std{5}   & 0\std{0}   & 0\std{0}    & 0\std{0}    & 0\std{0}            & 6\std{5}           & 94\std{3}         & \best{97\std{3}} \\
        & \texttt{giant-navigate}   & 65\std{5}  & 0\std{0}   & 14\std{3}   & 16\std{3}   & 1\std{1}            & 73\std{5}   & 78\std{3}         & \best{88\std{2}} \\
        & \texttt{giant-stitch}     & 2\std{2}   & 0\std{0}   & 0\std{0}    & 0\std{0}    & 3\std{1}   & 36\std{7}   & \best{88\std{4}}  & \best{88\std{2}} \\
        & \texttt{teleport-navigate}& 42\std{3}  & 35\std{5}  & 35\std{5}   & 53\std{2}   & 27\std{12}           & 29\std{13}          & 39\std{3}           & \best{90\std{1}} \\
        & \texttt{teleport-stitch}  & 36\std{2}  & 17\std{2}  & 17\std{2}   & 24\std{5}   & 29\std{13}           & 36\std{14}          & 39\std{3}           & \best{92\std{2}} \\
        & \texttt{teleport-explore} & 34\std{2}  & 7\std{3}   & 2\std{2}    & 20\std{2}   & 50\std{2}  & 53\std{8}           & 32\std{2}           & \best{93\std{2}} \\
        \cmidrule(lr){2-10}
        & \texttt{mean} & 43 & 12 & 20 & 26 & 29 & 50 & 70 & \best{91} \\
        \midrule
        \multirow{6}{*}{\texttt{pointmaze}}
        & \texttt{large-navigate}   & 58\std{5}  & 34\std{3}  & 86\std{9}   & 45\std{5}   & 34\std{12}           & 92\std{4} & 30\std{11}       & \best{96\std{5}} \\
        & \texttt{large-stitch}     & 13\std{6}  & 31\std{3}  & 84\std{15}  & 12\std{6}   & 23\std{8}            & \best{98\std{2}} & 63\std{20}        & \best{98\std{2}} \\
        & \texttt{giant-navigate}   & 46\std{9}  & 0\std{0}   & 68\std{7}   & 27\std{10}  & 1\std{1}            & 80\std{12}  & 3\std{6}                & \best{96\std{3}} \\
        & \texttt{giant-stitch}     & 0\std{0}   & 0\std{0}   & 50\std{8}   & 0\std{0}    & 0\std{0}            & 81\std{17}  & 0\std{0}                  & \best{99\std{1}} \\
        & \texttt{teleport-navigate}& 18\std{4}  & 24\std{7}  & 4\std{4}    & 45\std{3}   & 28\std{4}           & 28\std{4}          & 17\std{9}                 & \best{99\std{2}} \\
        & \texttt{teleport-stitch}  & 34\std{4}  & 25\std{3}  & 9\std{5}    & 44\std{2}   & 29\std{2}  & 37\std{26}          & 5.6\std{4}                  & \best{98\std{2}} \\
        \cmidrule(lr){2-10}
        & \texttt{mean} & 28 & 19 & 50 & 29 & 19 & 69 & 20 & \best{98} \\
        \midrule
        \multirow{6}{*}{\texttt{visual-antmaze}}
        & \texttt{medium-navigate}  & 93\std{4}  & 11\std{1}  & 0\std{0}  & 94\std{1}  & 9\std{2}  & 71\std{5} & 96\std{1}         & \best{97\std{1}} \\
        & \texttt{medium-stitch}    & 87\std{2}  & 2\std{0}   & 0\std{0}  & 69\std{2}  & 63\std{2} & 30\std{4} & 90\std{3}         & \best{96\std{1}} \\
        & \texttt{medium-explore}   & 0\std{0}   & 0\std{0}   & 0\std{0}  & 0\std{0}   & 0\std{0}  & 38\std{10} & 66\std{7}         & \best{84\std{4}} \\
        & \texttt{large-navigate}   & 53\std{9}  & 4\std{1}   & 0\std{0}  & 84\std{1}  & 3\std{1}  & 37\std{12} & \best{87\std{1}} & 80\std{4} \\
        & \texttt{large-stitch}     & 28\std{2}  & 0\std{0}   & 1\std{1}  & 11\std{3}  & 18\std{5} & 7\std{2} & 75\std{4}         & \best{88\std{1}} \\
        & \texttt{giant-stitch}     & 0\std{0}   & 0\std{0}   & 0\std{0}  & 0\std{0}   & 1\std{1}  & 2\std{2} & 56\std{4}         & \best{72\std{5}} \\
        \cmidrule(lr){2-10}
        & \texttt{mean} & 44 & 3 & 0 & 43 & 16 & 31 & 78 & \best{86} \\
        \bottomrule
    \end{tabular*}
    \vspace{-0.5em}
\end{table*}

\subsection{Performance Comparison}

Table~\ref{tab:ogbench_performance} summarizes the success rates across all evaluated environments. Overall, BGP consistently outperforms prior offline GCRL methods, achieving mean success rates of 91\% on \texttt{antmaze}, 98\% on \texttt{pointmaze}, and 86\% on \texttt{visual-antmaze}, substantially exceeding our primary baseline, HIQL and demonstrating the effectiveness of branching-aware graph-based planning under substantial stochasticity. This advantage is particularly evident in large-scale layouts such as \texttt{large} and \texttt{giant}, where many prior methods achieve only partial success across goals, whereas BGP reliably reaches most targets. Methods that incorporate graph structure or trajectory stitching, such as GAS and CGCIVL, also improve in these settings, although their performance varies across environments and dataset types, for example on \texttt{pointmaze} and the \texttt{explore} datasets. Overall, BGP maintains strong performance across layouts and datasets, indicating robust behavior under diverse conditions.

The advantage becomes more pronounced in the \texttt{teleport} environments, where stochastic transitions create deceptive branches that can mislead planning. Most prior methods achieve limited success, and even approaches such as TMD, which explicitly address these challenges, achieve only 20--50\% success rates. In contrast, BGP consistently performs well across most goals, with remaining failures primarily attributable to minor control errors rather than planning limitations. These results highlight the importance of explicitly modeling stochastic branching. By identifying branching structures through landmarks, particularly near teleport regions, and avoiding risky branches leading to dead zones, as illustrated in Fig.~\ref{fig:figure4}(c), BGP enables more reliable long-horizon planning under complex dynamics. Appendix~\ref{subsecapp:additional_tasks} further reports substantial improvements over the compared baselines on \texttt{humanoidmaze} and \texttt{manipulation} tasks.

\subsection{Path Analysis under Graph Planning}

\begin{figure}[!t]
    \centering
    \includegraphics[width=\linewidth]{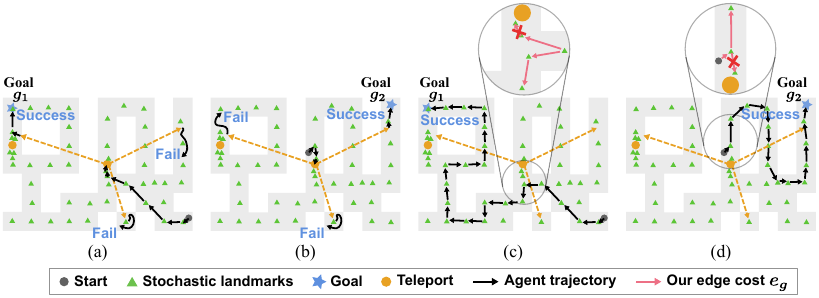}
    \caption{Shortest waypoint paths in the \texttt{teleport} layout. (a–b) use the prior edge cost $e_p(v,w)=-V_\theta(v,w)$ and illustrate failure cases, while (c–d) use the proposed goal-conditioned edge cost $e_g$. Goals are selected from the evaluation set. (c–d) further illustrate how $e_g$ accounts for downstream branches and avoids unreliable teleport transitions, yielding more stable paths.}
    \vspace{-1em}
    \label{fig:graph construction}
\end{figure}

To understand how BGP handles stochastic branching, we visualize the learned landmarks and resulting shortest paths under different edge constructions. Fig.~\ref{fig:graph construction} presents representative results on the \texttt{teleport} layout, where performance differences are most pronounced, while Appendix~\ref{subsecapp:trajectory_analysis} provides full results across all layouts and goals. We fix the same landmark set and compare paths under the prior edge $e_p(v,w)=-V_\theta(v,w)$ and the proposed goal-conditioned edge $e_g$.

The learned landmarks capture stochastic branching structure across goals, including teleport regions and other divergence points, but the choice of edge cost critically affects planning. As shown in Fig.~\ref{fig:graph construction}(a) and (b), with $e_p$, all locally reachable transitions are treated as valid, so the planner favors seemingly short teleport paths that often fail, resulting in a \emph{path mirage} with success rates around 33\%. In contrast, Fig.~\ref{fig:graph construction}(c) and (d) show that $e_g$ uses downstream reachability to disconnect branches leading to dead zones, preserve feasible ones, and consistently select stable paths.

Notably, BGP selects paths based on the goal rather than simply eliminating risky branches: teleport transitions are retained when needed for goal reaching and avoided otherwise. This generalizes beyond teleport-specific cases: as shown in Appendix~\ref{subsecapp:trajectory_analysis}, in \texttt{large} and \texttt{giant} layouts, where no explicit dead zones exist, edges remain connected and planning reduces to selecting reliable paths among many alternatives. Overall, \emph{BGP effectively handles branching-induced stochasticity and enables stable planning} across both highly branching and general long-horizon settings.

\subsection{Robustness to Increasing Stochasticity}

\begin{figure}[!t]
    \centering
    \includegraphics[width=\linewidth]{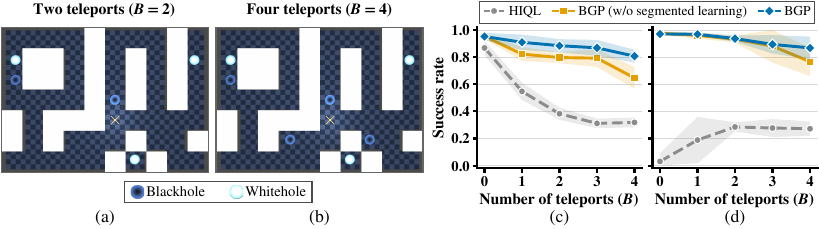}
\caption{Environments and success rates with respect to the number of teleports $B$. Panels (a) and (b) show the environments with two and four teleports, respectively. Panel (c) shows the success rates of HIQL, BGP (w/o segmented learning), and BGP on the \texttt{navigate} datasets, and panel (d) shows those on the \texttt{stitch} datasets. Shaded regions denote 95\% confidence intervals over 5 seeds.}
\label{fig:teleport_k}
    \vspace{-1em}
\end{figure}

We further investigate the robustness of BGP under increasing stochasticity by varying the number of active teleport entrances, referred to as blackholes, $B \in \{0,\dots,4\}$ in \texttt{antmaze-teleport}, where $B{=}2$ is the original environment, as illustrated in Fig.~\ref{fig:teleport_k}(a) and (b) for $B{=}2$ and $B{=}4$. We evaluate BGP with and without segmented learning to examine its contribution to robustness. For each $B$, we collect \texttt{navigate} and \texttt{stitch} datasets with the same random seed following the OGBench protocols, so that they differ only in environment dynamics, as detailed in Appendix~\ref{app:teleport_k_data}.

\textbf{Navigate.} As shown in Fig.~\ref{fig:teleport_k}(c), HIQL achieves a high success rate without teleports, but its performance drops substantially as $B$ increases. In contrast, both BGP variants maintain high success rates with little degradation, demonstrating robustness to increasing stochasticity. The gap between the two variants also widens as $B$ increases, suggesting that segmented learning further improves robustness by enabling more reliable value estimation over shorter segments.

\textbf{Stitch.} Since the \texttt{stitch} datasets consist of short episodes, learning reliable values for distant goals is challenging. As shown in Fig.~\ref{fig:teleport_k}(d), HIQL fails to reach the goal even without teleports due to unstable value estimation. With teleports, HIQL instead relies on chance transitions to reach the goal, achieving approximately 33\% success. In contrast, both BGP variants maintain high success rates, performing similarly for smaller $B$. At $B{=}4$, BGP achieves approximately 90\% success, outperforming BGP (w/o segmented learning) by about 10\% points. These results highlight \emph{BGP's robustness to increasing stochasticity} and \emph{the benefit of segmented learning} for reliable value estimation over shorter segments.

\vspace{9pt}
\subsection{Further Analysis of BGP}

\begin{figure}[!t]
\centering

\begin{minipage}[b]{0.41\linewidth}
\centering

\resizebox{\linewidth}{!}{%
\vspace{1em}
\begin{tabular}{lc}
    \toprule
    \textbf{Variant} & \textbf{Success Rate (\%)} \\
    \midrule
    BGP (Default) & $\mathbf{91.2 \pm 1.3}$ \\
    \midrule
    w/ one-step estimation  & $86.7 \pm 1.3$ \\
    w/o segmented learning & $84.4 \pm 6.2$ \\
    w/ random landmarks & $69.4 \pm 4.2$ \\
    w/ prior edge & $44.3 \pm 3.6$ \\
    w/ value-based distance & $46.8 \pm 5.2$ \\
    \bottomrule
\end{tabular}%
}

\vspace{1.5em}
\captionof{table}{Component evaluation.}
\label{tab:component_ablation}

\end{minipage}\hfill
\begin{minipage}[b]{0.57\linewidth}
\centering

\vspace{-0.4em}
\begin{subfigure}[b]{0.48\linewidth}
    \centering
    \includegraphics[width=\linewidth]{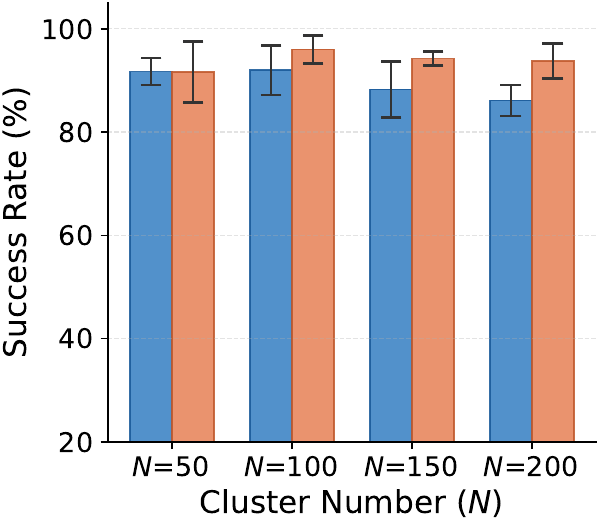}
    \captionsetup{skip=2pt}
    \caption{}
    \label{fig:sensitivity_ncluster}
\end{subfigure}
\hfill
\begin{subfigure}[b]{0.48\linewidth}
    \centering
    \includegraphics[width=\linewidth]{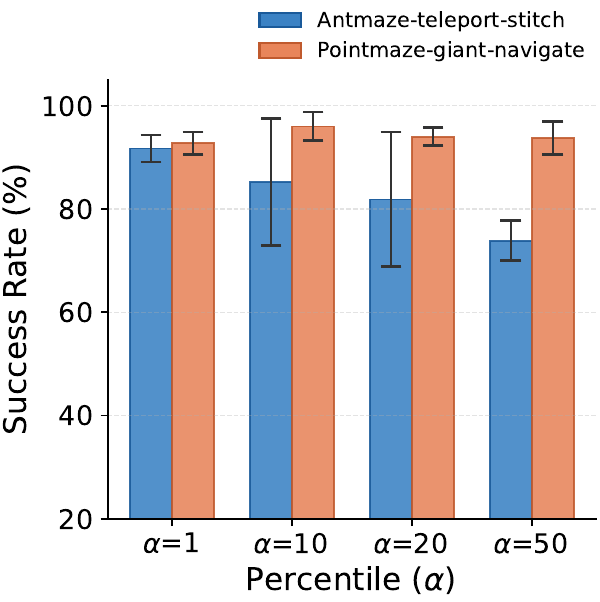}
    \captionsetup{skip=2pt}
    \caption{}
    \label{fig:sensitivity_percentile}
\end{subfigure}

\vspace{0pt}
\captionof{figure}{Hyperparameter analysis.}
\label{fig:hyperparameter_ablation}

\end{minipage}

\end{figure}

\paragraph{Component evaluation.}
We evaluate the contribution of each component in BGP through a set of ablations: \textbf{BGP w/ one-step estimation}, which uses a one-step predictor ($n=1$) instead of the Transformer-based uncertainty model; \textbf{w/ random landmarks}, which uniformly samples landmarks from the dataset without uncertainty guidance; \textbf{w/o segmented learning}, which trains on full transitions; \textbf{w/ prior edge}, which replaces $e_g$ with $e_p=-V_\theta$; and \textbf{w/ value-based distance}, which directly uses $d(s,g)=-V_\theta(s,g)$ without constructing the pre-graph $\mathcal{G}_p$. Table~\ref{tab:component_ablation} reports the average success rate across all variants on \texttt{antmaze-teleport-navigate,stitch}. BGP (Default) consistently outperforms all variants, indicating that each component contributes to the final performance. One-step estimation and random landmark selection degrade performance, highlighting the importance of accurately capturing uncertainty to construct informative landmarks. Most critically, both the prior edge and value-based distance variants cause severe performance drops, confirming that edge cost design is the dominant factor in BGP, consistent with the path analysis.

\paragraph{Effect of the number of clusters $N$.}
We analyze the effect of the number of landmark clusters $N$ in Fig.~\ref{fig:hyperparameter_ablation}(a), sweeping $N \in \{50, 100, 150, 200\}$ on \texttt{antmaze-teleport-stitch} and \texttt{pointmaze-giant-navigate}. In \texttt{pointmaze-giant-navigate}, performance is largely stable across values of $N$, suggesting that once meaningful landmarks are identified, navigation is straightforward. In contrast, in \texttt{antmaze-teleport-stitch}, performance slightly degrades as $N$ increases, since larger $N$ induces overly dense local connectivity and increases planning instability. However, the variation is moderate, indicating low sensitivity to $N$ within a reasonable range. In practice, $N \in [50,100]$ provides a good balance between coverage and stability.

\paragraph{Effect of the percentile threshold $\alpha$.}
We analyze the percentile threshold $\alpha$ in Fig.~\ref{fig:hyperparameter_ablation}(b), sweeping $\alpha \in \{1, 10, 20, 50\}$, which determines which high-uncertainty states are selected as landmarks. Sensitivity to $\alpha$ depends on how stochastic branching is distributed. In \texttt{pointmaze-giant-navigate}, performance remains stable across values, as branching is relatively diffuse. In contrast, \texttt{antmaze-teleport-stitch} is highly sensitive: performance drops as $\alpha$ increases. Large $\alpha$ values bias landmark selection toward state diversity rather than true branching structure, resulting in graphs that fail to capture critical transitions. Consequently, the planner cannot reliably avoid failure-prone paths. In practice, smaller $\alpha$ values are preferable in environments with concentrated branching.

\paragraph{Additional analysis.}
We provide additional hyperparameter analyses across a broader range of layouts and settings in Appendix~\ref{subsecapp:hyperparameter_sensitivity}. Following \citet{jackson2026clean}, we also conduct a budget-aware hyperparameter search in Appendix~\ref{subsecapp:budget_aware}, demonstrating that BGP achieves strong performance with a modest search budget. Finally, we analyze computational costs in Appendix~\ref{subsecapp:compute_cost}, showing that the additional overhead introduced by BGP remains modest in practice.

\vspace{-0.5em}
\section{Limitation}
\label{sec:limitation}

While BGP demonstrates strong performance, several limitations remain. First, BGP introduces additional hyperparameters, including the percentile threshold and the number of clusters. However, our analysis in the Appendix evaluates both hyperparameter sensitivity and search costs, providing practical guidance for efficient tuning. Second, graph construction and planning introduce additional computational overhead. Our cost analysis in the Appendix shows that this overhead is modest relative to the overall training cost. Finally, while BGP remains effective in high-dimensional manipulation tasks, more sophisticated state representations may further improve its performance in such settings. We leave this direction for future work.

\vspace{-0.5em}
\section{Conclusion}
\label{sec:conclusion}

We presented BGP, a branch-aware framework for offline goal-conditioned reinforcement learning that addresses unreliable planning under branching-induced stochasticity in challenging long-horizon tasks. By combining uncertainty-guided landmark construction with goal-conditioned graph planning, BGP effectively captures critical branching structures and enables stable waypoint planning through its two-phase graph structure. Segmented learning further supports reliable local value estimation. Extensive experiments on OGBench demonstrate consistent improvements over existing offline GCRL methods, particularly in challenging long-horizon tasks, highlighting the effectiveness of BGP for reliable goal-conditioned control.
\newpage

\section*{Acknowledgements}
This work was partly supported by the Institute of Information \& Communications Technology Planning \& Evaluation (IITP) grant funded by the Korea government (MSIT)  (No. RS-2020-
II201336, Artificial Intelligence Graduate School Support (UNIST)) and (No. RS-2025-25442824, AI Star-Fellowship Program (UNIST)), and partly by the National Research Foundation of Korea (NRF) grants funded by the Korea government (MSIT) (No. RS-2025-23523191, LLM-Based Multi-Agent Reinforcement Learning for End-to-End Large Autonomous Swarm Control) and (No. RS-2026-25536992, AI-Based Nuclear Fusion Heating Technology and Digital Twin Development via Synthetic Diagnostics).

\bibliographystyle{unsrtnat}
\bibliography{neurips_2026}

\newpage
\appendix

\section{Implementation Details}
\label{secapp:implementation_details}
This section presents additional implementation details of the proposed framework. The branch-aware graph construction process is detailed in Appendix~\ref{subsecapp:graph_construction}, and the offline learning procedure combined with the segmented learning scheme is described in Appendix~\ref{subsecapp:segmented_learning}.
 
\subsection{Graph Construction Detail}
\label{subsecapp:graph_construction}
 
\begin{figure}[h]
    \centering
    \includegraphics[width=0.7\textwidth]{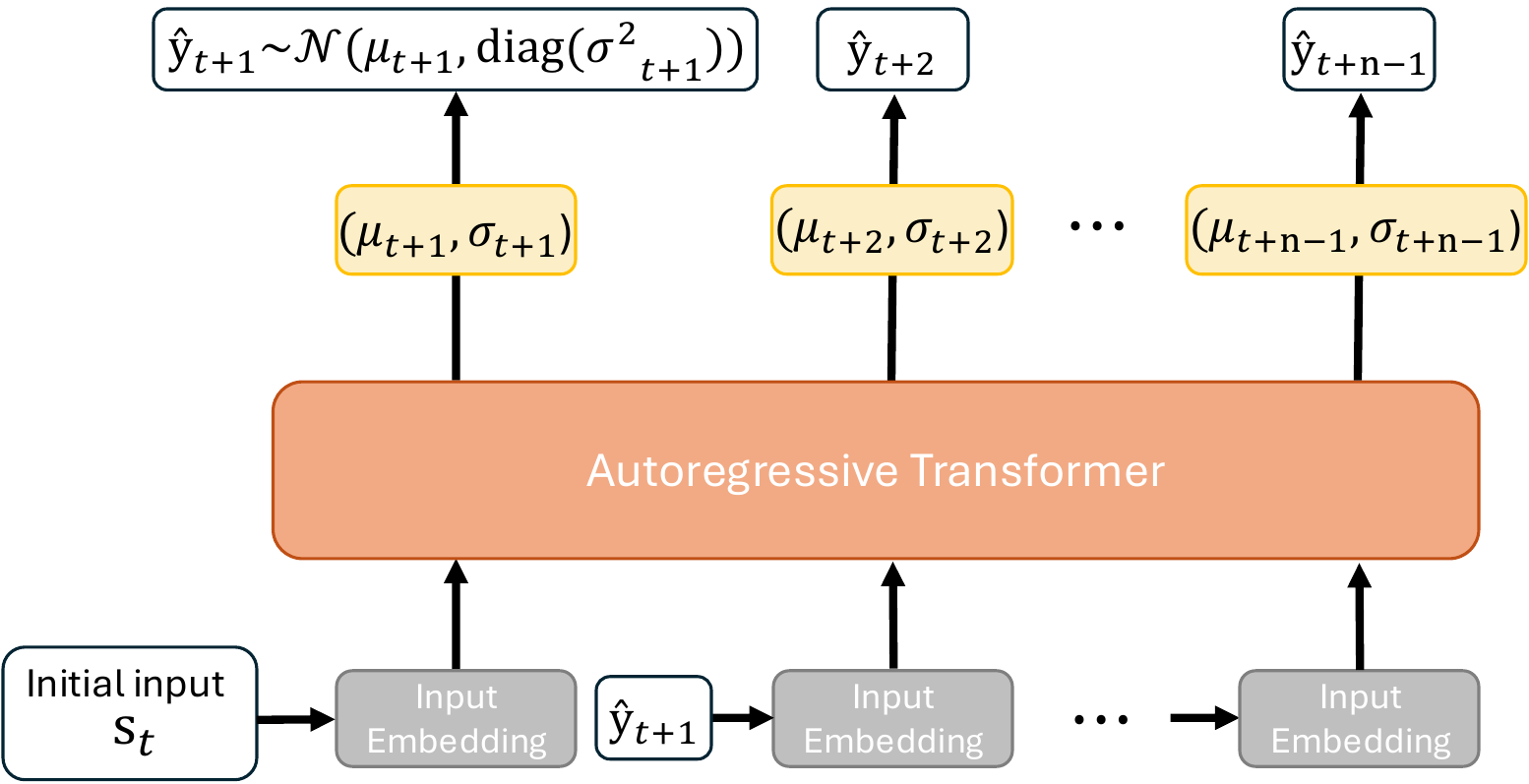}
    \caption{Autoregressive prediction of the Transformer during evaluation, where predicted transition variables $\hat{y}_{t+l}$ are fed back as inputs to estimate multi-step uncertainty from the current state $s_t$.}
    \label{fig:transformer_arch}
\end{figure}

\textbf{Uncertainty model.} To compute the uncertainty score $\mathcal{U}^{(n)}$ in Eq.~(\ref{eq:uncertainty_score}), we parameterize the autoregressive model $p_\psi$ as a decoder-only Transformer that predicts future joint transition variables $y_{t+l} = (a_{t+l}, s_{t+l+1})$. Rather than predicting deterministic targets, the Transformer outputs the parameters $(\mu_{t+l}, \sigma_{t+l})$ of a conditional Gaussian, $p_\psi(y_{t+l} \mid s_t, y_{t:t+l-1}) = \mathcal{N}\bigl(y_{t+l}; \mu_{t+l}, \mathrm{diag}(\sigma_{t+l}^2)\bigr)$, so that the entropy in $\mathcal{U}^{(n)}$ can be computed directly from the predicted variance $\sigma_{t+l}$. The model is trained with teacher forcing on input sequences $[s_t, y_t, \ldots, y_{t+n-1}]$, where the output at each position predicts the next token under causal masking, by minimizing the negative log-likelihood over $\mathcal{D}$:
\begin{equation}
\mathcal{L}(\psi) = \mathbb{E}_{(s_t, y_{t:t+n-1}) \sim \mathcal{D}} \Bigl[ \textstyle\sum_{l=0}^{n-1} - \log p_\psi \bigl(y_{t+l} \mid s_t, y_{t:t+l-1}\bigr) \Bigr].
\end{equation}
At evaluation, the predictions are fed back into the Transformer autoregressively, as illustrated in Fig.~\ref{fig:transformer_arch}, and $\mathcal{U}^{(n)}(s_t)$ is computed as the discounted sum of the resulting conditional entropies over $n$ future steps.
 
\textbf{Representation and clustering.} To handle high-dimensional inputs, we project states into a representation space $\mathcal{Z}$ before clustering. For state-based \texttt{antmaze} and \texttt{pointmaze}, we use a 2-dimensional PCA projection~\citep{abdi2010principal}. For the other environments, we use a Hilbert representation pretrained with goal-conditioned IQL on an IMPALA-small CNN encoder following~\citet{park2024foundation}, which is then frozen. For visual tasks, the Transformer is trained and evaluated on top of the frozen representation. In this space, the top-$\alpha$ percentile high-uncertainty states are clustered with K-means++, and each landmark is chosen as the actual dataset state closest to its cluster center, so that every graph vertex corresponds to an actual state near a stochastic branching point. Table~\ref{tab:environment_representations} summarizes the representation and its dimension used in each environment, including \texttt{humanoidmaze} and the manipulation tasks evaluated in the additional experiments in Appendix~\ref{subsecapp:additional_tasks}.
 
\textbf{Planning.} Given a goal, the goal-conditioned edge cost $e_g$ in Eq.~(\ref{eq:edge}) is computed from the expected distance reduction toward the goal. Since $e_g$ becomes negative when a transition reliably reduces the distance to the goal, we add a constant shift to all valid edges so that all weights are non-negative, as required by Dijkstra's algorithm, before computing the shortest waypoint path.
 
\begin{table}[ht]
    \centering
    \small
    \renewcommand{\arraystretch}{1.2}
    \caption{Representations used for landmark clustering in each environment. Dimension denotes the dimensionality of the representation space $\mathcal{Z}$.}
    \begin{tabular}{llc}
        \toprule
        \textbf{Environment} & \textbf{Representation for clustering} & \textbf{Dimension} \\
        \midrule
        \texttt{antmaze}        & PCA projection & 2 \\
        \texttt{pointmaze}      & PCA projection & 2 \\
        \texttt{visual-antmaze} & Hilbert representation & 16 \\
        \texttt{humanoidmaze}   & Hilbert representation & 16 \\
        \texttt{manipulation tasks}      & Hilbert representation & 16 \\
        \bottomrule
    \end{tabular}
    \label{tab:environment_representations}
\end{table}

\newpage

\subsection{Offline Learning Detail with Segmented Learning}
\label{subsecapp:segmented_learning}
 
\begin{figure}[h]
    \centering
    \includegraphics[width=0.8\textwidth]{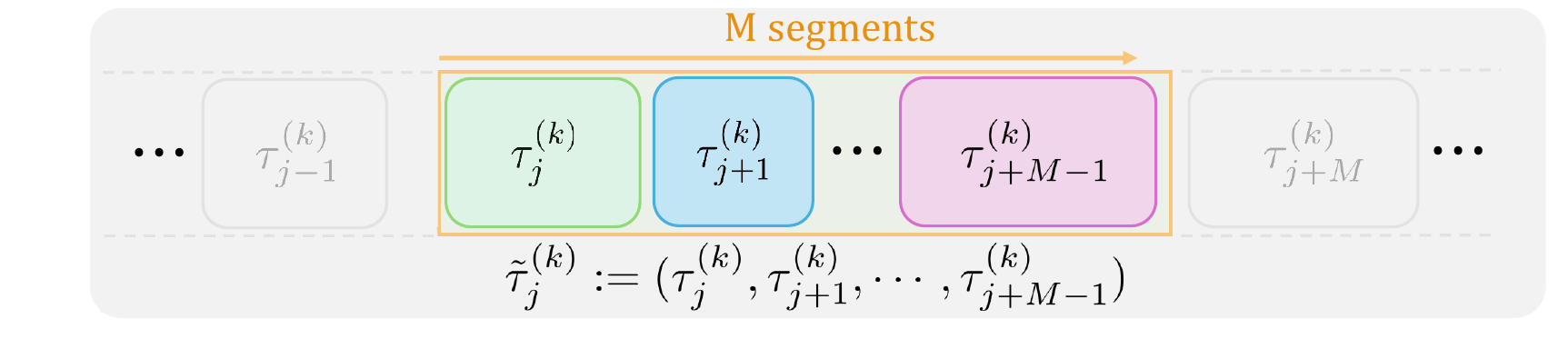}
    \caption{Illustration of segmented learning. Each offline trajectory is partitioned into sub-trajectories $\tau_j^{(k)}$ at cluster transition boundaries, shown in different colors, and $M$ consecutive sub-trajectories are concatenated into a local training window $\tilde{\tau}_j^{(k)}$.}
    \label{fig:segmented_learning}
\end{figure}

As described in Section~\ref{sec:segmented Learning} and illustrated in Fig.~\ref{fig:segmented_learning}, we train the value function and policy on local windows $\tilde{\tau}_j^{(k)}$, each formed by concatenating $M$ consecutive sub-trajectories of the $k$-th offline trajectory. The choice of $M$ balances local trajectory coverage against learning stability. A small $M$ may miss useful skip-like transitions and longer local transitions, while a large $M$ includes many stochastic boundaries, which leads to accumulated stochasticity and value-estimation bias. Within each window, the goal for a given state is sampled from the current state, a future state in the same window, or a random state in the window, with the ratios listed in Table~\ref{tab:fixed_hparams}.
 
We use HIQL as the offline reinforcement learning backbone. At execution, given the waypoint path $(\tilde{w}_0, \ldots, \tilde{w}_{K'})$ computed by the branch-aware graph planner, the learned policy sequentially aims to reach each waypoint $\tilde{w}_i$. The high-level policy $\pi^h(s_g \mid s, \tilde{w}_i)$ decomposes $\tilde{w}_i$ into a closer subgoal $s_g$, and the low-level policy $\pi^l(a \mid s, s_g)$ interacts with the environment to reach it. The objective function of HIQL is provided in Appendix~\ref{subsecapp:other_baseline}.

\newpage

\subsection{Environmental Details}
\label{subsecapp:environment_detail}

\begin{figure}[h]
    \centering
    \includegraphics[width=0.9\textwidth]{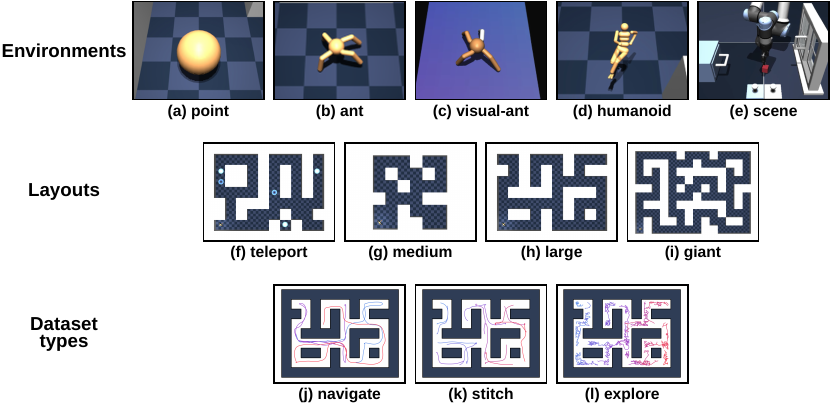}
    \caption{Visualizations of the OGBench environments used in our experiments. Panels (a)-(e) show the agents and the manipulation environment, panels (f)-(i) show the maze layouts, and panels (j)-(l) show example trajectories of the three dataset types.}
    \label{fig:app_environments}
\end{figure}

We conducted all experiments on an NVIDIA RTX 3090 GPU and an Intel Xeon Gold 6348 CPU running Ubuntu 20.04. We evaluate our method on OGBench~\citep{park2024ogbench}, a standard benchmark designed for offline goal-conditioned reinforcement learning. We specifically focus on environments where the offline data exhibit substantial stochasticity originating from both complex environment dynamics and suboptimal behavior policies.

\paragraph{Environments.}All continuous control environments are simulated within the MuJoCo physics engine~\citep{todorov2012mujoco}.
\begin{itemize}[leftmargin=*, topsep=0pt]
    \item \texttt{pointmaze:} As shown in Fig.~\ref{fig:app_environments}(a), the agent governs a 2D point mass via a 2-DoF action space. The observation space is 2-dimensional, consisting of the agent's position. While executing low-level actions is simpler compared to \texttt{antmaze}, the high-level routing complexity remains identical, isolating the challenge of long-horizon planning.
    \item \texttt{antmaze:} As shown in Fig.~\ref{fig:app_environments}(b), the agent operates a quadruped Ant robot equipped with 8 degrees of freedom (DoF). Solving this environment requires the agent to seamlessly integrate high-level topological planning with low-level complex locomotion. The 29-dimensional observation space includes the agent's absolute 2D $(x, y)$ position alongside joint kinematics.
    \item \texttt{visual-antmaze:} As shown in Fig.~\ref{fig:app_environments}(c), this environment uses the same Ant robot and maze layouts as \texttt{antmaze}, but the agent receives image observations instead of states, which additionally requires learning from high-dimensional pixel inputs.
    \item \texttt{humanoidmaze:} As shown in Fig.~\ref{fig:app_environments}(d), this environment uses the \texttt{medium}, \texttt{large}, and \texttt{giant} maze layouts, where the agent controls a much more complex 21-DoF Humanoid robot. As in other maze environments, the agent must learn both high-level maze navigation and low-level locomotion skills, and reaching distant goals in the \texttt{giant} layout requires up to 3,000 environment steps.
    \item \texttt{manipulation tasks:} As shown in Fig.~\ref{fig:app_environments}(e), among the manipulation environments in OGBench, we use \texttt{scene}, which requires long-horizon sequential manipulation. A 6-DoF UR5e robot arm interacts with a cube, a window, a drawer, and two button locks, and we use the \texttt{play} dataset collected by scripted policies that randomly interact with the objects.
\end{itemize}

\paragraph{Maze layouts.}
To rigorously test our stochasticity-aware framework, we utilize four maze layouts provided by OGBench:
\begin{itemize}[leftmargin=*, topsep=0pt]
    \item \texttt{teleport}: As shown in Fig.~\ref{fig:app_environments}(f), this layout introduces severe environmental stochasticity through unpredictable state transitions caused by blackholes and whiteholes. It serves as a critical testbed for evaluating our method's ability to resist lucky sample bias and construct robust waypoint graphs.
    \item \texttt{medium}: As shown in Fig.~\ref{fig:app_environments}(g), this layout follows the original medium maze in D4RL and provides a shorter task horizon than the other layouts.
    \item \texttt{large} and \texttt{giant}: As shown in Fig.~\ref{fig:app_environments}(h) and (i), these highly scaled layouts significantly extend the task horizon. They are specifically designed to test the agent's vulnerability to accumulated over-optimism and value estimation errors over extended distances.
\end{itemize}

\paragraph{Dataset types and evaluation setup.}
To assess algorithm performance across diverse data distributions, we utilize three types of offline datasets:
\begin{itemize}[leftmargin=*, topsep=0pt]
\item \textbf{\texttt{navigate}:} As shown in Fig.~\ref{fig:app_environments}(j), this dataset serves as the standard benchmark setting, where trajectories are generated by a goal-directed policy with injected noise. It evaluates the agent's ability to perform general goal-conditioned navigation.
\item \textbf{\texttt{stitch}:} As shown in Fig.~\ref{fig:app_environments}(k), this dataset is composed of short, disconnected trajectory segments that do not individually solve the task. The agent must combine multiple segments to form a valid long-horizon solution, emphasizing its ability to reason over fragmented experience.
\item \textbf{\texttt{explore}:} As shown in Fig.~\ref{fig:app_environments}(l), this dataset contains highly noisy and unstructured trajectories with broad state coverage but low data quality. It evaluates whether the agent can extract useful navigation behavior from poorly structured experience.
\end{itemize}

\newpage

\subsection{Details of Baselines and Backbones}
\label{subsecapp:other_baseline}
\begin{itemize}[leftmargin=*]

    \item \textbf{GCIQL}~\citep{park2023hiql} is a goal-conditioned implicit value-learning method that learns both a state-value function and an action-value function under the IQL framework:

\begin{equation}
\mathcal{L}_V(\theta)
=
\mathbb{E}_{(s,a)\sim \mathcal{D}}
\left[
\ell_\kappa^2\left(
\bar{Q}_{\bar{\theta}}(s,a,g)-V_{\theta}(s,g)
\right)
\right],
\label{eq:gciql_value}
\end{equation}

\begin{equation}
\mathcal{L}_Q(\theta)
=
\mathbb{E}_{(s,a,s')\sim \mathcal{D}}
\left[
\left(
r(s,g)+\gamma V_{\bar{\theta}}(s',g)-Q_{\theta}(s,a,g)
\right)^2
\right],
\label{eq:gciql_q}
\end{equation}

During training, the goal $g$ is sampled from a mixture of the current state, a future state from the same trajectory as $s$, and a random state from the full offline dataset.

To extract a policy from these learned value functions, GCIQL can use either a value-only AWR objective or a DDPG+BC objective. In our experiments, we mainly use the DDPG+BC objective, which has been reported to outperform AWR on OGBench~\citep{park2024ogbench}. During actor training, the goal $g$ is sampled only from future states in the same trajectory as $s$, resulting in the following objective:
\begin{equation}
\mathcal{L}_{\pi}(\psi)
=
\mathbb{E}_{(s,a)\sim \mathcal{D}}
\left[
Q_{\theta}\left(s,\pi_{\psi}(s,g),g\right)
+
\alpha \log \pi_{\psi}(a\mid s,g)
\right],
\label{eq:gciql_policy}
\end{equation}

Open-source code of GCIQL is available at
\href{https://github.com/seohongpark/ogbench}
{https://github.com/seohongpark/ogbench}.

\item \textbf{HIQL}~\citep{park2023hiql} is a hierarchical goal-conditioned reinforcement learning method that extends implicit value learning to long-horizon tasks by decomposing control into a high-level subgoal policy and a low-level goal-reaching policy over subgoals. The goal-conditioned value function is trained via expectile regression:

\begin{equation}
\mathcal{L}_V(\theta)
=
\mathbb{E}_{(s,s')\sim \mathcal{D}}
\left[
\ell_\kappa^2\left(
r(s,g)+\gamma V_{\bar{\theta}}(s',g)-V_{\theta}(s,g)
\right)
\right],
\label{eq:hiql_value2}
\end{equation}

During value-function training, the goal $g$ is sampled from a mixture of the current state, a future state from the same trajectory as $s$, and a random state from the full offline dataset.
HIQL then learns the high-level policy $\pi^h_{\psi_h}(s_{t+k}\mid s_t,g)$ and the low-level policy $\pi^l_{\psi_l}(a\mid s_t,s_{t+k})$ via advantage-weighted regression:

\begin{equation}
\mathcal{L}_{\pi^h}(\psi_h)
=
\mathbb{E}_{(s_t,s_{t+k})\sim \mathcal{D}}
\left[
\exp\left(
\beta \tilde{A}^h(s_t,s_{t+k},g)
\right)
\log \pi^h_{\psi_h}(s_{t+k}\mid s_t,g)
\right],
\label{eq:hiql_high}
\end{equation}

\begin{equation}
\mathcal{L}_{\pi^l}(\psi_l)
=
\mathbb{E}_{(s_t,a_t,s_{t+1},s_{t+k})\sim \mathcal{D}}
\left[
\exp\left(
\beta \tilde{A}^{l}(s_t,a_t,s_{t+k})
\right)
\log \pi^l_{\psi_l}(a_t\mid s_t,s_{t+k})
\right],
\label{eq:hiql_low}
\end{equation}

During actor training, the goal $g$ is sampled only from future states in the same trajectory as $s$. Here, $\beta$ denotes a temperature hyperparameter for advantage-weighted regression, and the advantages are approximated as
$\tilde{A}^h(s_t,s_{t+k},g)\approx V_{\theta}(s_{t+k},g)-V_{\theta}(s_t,g)$
and
$\tilde{A}^{l}(s_t,a_t,s_{t+k})\approx V_{\theta}(s_{t+1},s_{t+k})-V_{\theta}(s_t,s_{t+k})$.
Open-source code of HIQL is available at
\href{https://github.com/seohongpark/ogbench}
{https://github.com/seohongpark/ogbench}.

    \item \textbf{QRL}~\citep{wang2023optimal} imposes a triangle inequality constraint using a quasimetric network architecture. This structural inductive bias allows the model to capture optimal asymmetric reachability and implicitly stitch shortest paths from suboptimal offline data. Open-source code of QRL is available at \href{https://github.com/seohongpark/ogbench}{https://github.com/seohongpark/ogbench}.
    
    \item \textbf{CRL}~\citep{eysenbach2022contrastive} formulates goal-conditioned RL as a representation learning problem. It utilizes a contrastive learning objective to estimate transition probabilities and temporal distances, avoiding the compounding errors typical of standard temporal difference learning. Open-source code of CRL is available at \href{https://github.com/seohongpark/ogbench}{https://github.com/seohongpark/ogbench}.
    
    \item \textbf{GAS}~\citep{baek2025graph} constructs explicit state graphs directly from static offline datasets. It performs global graph-based search to stitch disconnected trajectory segments, enabling the agent to find optimal paths without relying solely on complex policy network approximations. Open-source code of GAS is available at \href{https://github.com/qortmdgh4141/GAS}{https://github.com/qortmdgh4141/GAS}.
    
    \item \textbf{TMD}~\citep{myers2025offline} proposes Temporal Metric Distillation, which combines contrastive representation learning with quasimetric architectural constraints. By enforcing action invariance and the triangle inequality, it learns optimal goal-reaching distances even in highly stochastic offline environments. Open-source code of TMD is available at \href{https://github.com/vivekmyers/tmd-release/tree/master}{https://github.com/vivekmyers/tmd-release/tree/master}.

    \item \textbf{CGCIVL}~\citep{ke2025conservative} introduces a conservative value learning objective that explicitly penalizes unconnected state-goal pairs to prevent value overestimation during cross-trajectory sampling. By combining this conservative penalty with a quasimetric distillation mechanism, it ensures valid connected pairs maintain accurate distance estimates, effectively stabilizing long-horizon goal stitching. Open-source code of CGCIVL is available at \href{https://github.com/kkq2018/CGCIVL.git}{https://github.com/kkq2018/CGCIVL.git}.
\end{itemize}

\subsection{Hyperparameter Setup}
\label{subsecapp:hyperparameter_setup}
The hyperparameter settings for BGP are summarized in this section. Table~\ref{tab:fixed_hparams} lists the common hyperparameters shared across all environments, including network architectures and optimization details. We selected the Transformer context length $n=5$ and the local segment length $M=3$ as default values across all tasks because they provide a favorable balance between capturing sufficient multi-step stochasticity and maintaining computational efficiency during localized learning. Table~\ref{tab:best_hparams} details the environment-specific hyperparameters, specifically the top-$\alpha$ percentile for landmark identification and the number of clusters $N$, which are selected from the search ranges $\alpha \in \{1,10, 20, 50\}$ and $N \in \{50, 100, 150, 200\}$.

As noted in the main text, BGP relies on HIQL as the primary backbone and is trained for 500K steps on pixel-based tasks and 1M steps otherwise. For tasks that use the Hilbert representation, where clustering in the learned representation space is noisier, each local segment must contain at least 60 states, and a graph edge is retained only if the corresponding transition is observed at least 20 times in the dataset. Its performance remains robust across a reasonable range of these configurations. A detailed analysis of hyperparameter sensitivity is provided in Appendix~\ref{subsecapp:hyperparameter_sensitivity}.

\begin{table}[h]
\centering
\small
\setlength{\tabcolsep}{6pt}
\caption{Common hyperparameters.}
\label{tab:fixed_hparams}
\begin{tabular}{ll}
\toprule
\textbf{Hyperparameter} & \textbf{Value} \\
\midrule
\multicolumn{2}{c}{\textit{Transformer (Uncertainty Estimation)}} \\
\midrule
Number of layers & 4 \\
Number of attention heads & 2 \\
Embedding dimension & 256 \\
Context length ($n$) & 5 \\
Transformer learning rate & $3 \times 10^{-4}$ \\
\midrule
\multicolumn{2}{c}{\textit{Offline GCRL Backbone}} \\
\midrule
Segment length ($M$) & 3 \\
Batch size & 512(state-based task), 256(pixel-based task)\\
Value / Policy MLP dimensions & (512, 512, 512) \\
Nonlinearity & GELU \\
Value normalization & LayerNorm \\
Target network smoothing coefficient & 0.005 \\
Learning rate & 0.0003 \\
Optimizer & Adam \\
Discount factor $\gamma$ & 0.99(state-based task), 0.995(pixel-based task)\\
Subgoal selection frequency (HIQL) & 25(giantmaze, pixel-based task), 15(others)\\
$\epsilon$ for waypoint reachability & 1.5(state-based task), 40(pixel-based task) \\
Value $\left(p_{\text{cur}}^D, p_{\text{traj}}^D, p_{\text{rand}}^{\mathcal{D}}\right)$ ratio & (0.2, 0.5, 0.3) \\
Policy $\left(p_{\text{cur}}^D, p_{\text{traj}}^D, p_{\text{rand}}^{\mathcal{D}}\right)$ ratio & (0,1,0) (state-based task), (0, 0.5, 0.5) (pixel-based task) \\
\bottomrule
\end{tabular}
\end{table}

\begin{table}[ht]
\centering
\small
\setlength{\tabcolsep}{12pt}
\caption{Environment-specific hyperparameters.}
\label{tab:best_hparams}
\begin{tabular}{llcc}
\toprule
\textbf{Domain} & \textbf{Dataset} & \textbf{Percentile ($\alpha$)} & \textbf{Num Clusters ($N$)} \\
\midrule
\multirow{6}{*}{\texttt{pointmaze}}
& \texttt{teleport-navigate} & 1 & 100 \\
& \texttt{teleport-stitch}   & 1 & 100 \\
& \texttt{giant-navigate}    & 10 & 100 \\
& \texttt{giant-stitch}      & 10 & 100 \\
& \texttt{large-navigate}    & 1 & 50 \\
& \texttt{large-stitch}      & 1 & 50 \\
\midrule
\multirow{8}{*}{\texttt{antmaze}}
& \texttt{teleport-navigate} & 1 & 50 \\
& \texttt{teleport-stitch}   & 1 & 50 \\
& \texttt{teleport-explore}  & 10 & 200 \\
& \texttt{giant-navigate}    & 20 & 100 \\
& \texttt{giant-stitch}      & 20 & 100 \\
& \texttt{large-navigate}    & 10 & 50 \\
& \texttt{large-stitch}      & 10 & 50 \\
& \texttt{large-explore}     & 50 & 50 \\
\midrule
\multirow{6}{*}{\texttt{visual-antmaze}}
& \texttt{giant-stitch}      & 10 & 100 \\
& \texttt{large-navigate}    & 50 & 100 \\
& \texttt{large-stitch}      & 50 & 100 \\
& \texttt{medium-navigate}    & 10 & 100 \\
& \texttt{medium-stitch}      & 10 & 100 \\
& \texttt{medium-explore}     & 10 & 100 \\

\bottomrule
\end{tabular}
\end{table}

\newpage

\section{Additional Analysis and Ablation Studies}
\label{secapp:additional_analysis}

\subsection{Teleport Environments with Varying Numbers of Blackholes}
\label{app:teleport_k_data}

\begin{figure}[h]
    \centering
    \includegraphics[width=0.9\textwidth]{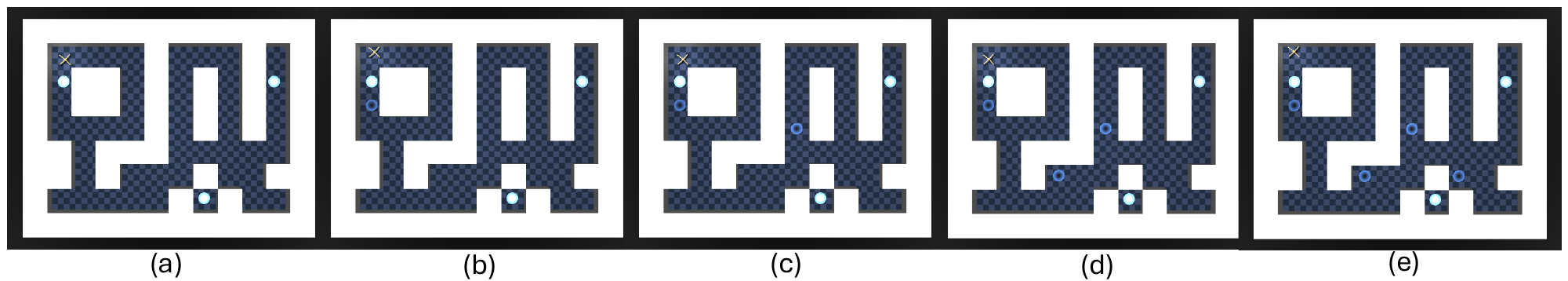}
\caption{\texttt{antmaze-teleport} environments with varying numbers of teleports. Panels (a)-(e) show the environments with $B{=}0$, $1$, $2$, $3$, and $4$ active blackholes, respectively, and panel (c) is the original OGBench environment.}
\label{fig:teleport_k_envs}
\end{figure}

\textbf{Environments.} Fig.~\ref{fig:teleport_k_envs} shows the five environments, which share the same maze layout and exit locations. When the agent enters an active blackhole, it is teleported to one of three fixed exits uniformly at random, with its velocity preserved.

\textbf{Offline datasets.} For each $B$, we collect \texttt{navigate} and \texttt{stitch} datasets following the OGBench protocols, where a pretrained SAC low-level policy with an action noise of 0.2 follows the subgoals of a shortest-path oracle that does not account for teleports. The \texttt{navigate} datasets consist of 1,000 training and 100 validation episodes of 1,001 steps, where a new goal is sampled whenever the current goal is reached. The \texttt{stitch} datasets consist of 5,000 training and 500 validation episodes of 201 steps, where each episode moves to a goal four cells away and stays near it.

\newpage

\subsection{Performance on Additional Tasks}
\label{subsecapp:additional_tasks}

To complement the main experiments, we provide additional results on humanoid locomotion and manipulation using \texttt{humanoidmaze} and \texttt{scene-play}, shown in Fig.~\ref{fig:app_environments}(d) and (e), respectively. For both environments, we perform landmark clustering in the Hilbert representation space $\mathcal{Z}$ following \citet{park2024foundation}.

\paragraph{Humanoidmaze.}
Table~\ref{tab:additional_tasks_performance} reports results on the \texttt{navigate} and \texttt{stitch} datasets of the \texttt{medium}, \texttt{large}, and \texttt{giant} layouts, together with the corresponding BGP hyperparameter settings. BGP achieves higher mean success rates than the compared baselines on five datasets, with particularly large improvements on the \texttt{giant} layouts.

\paragraph{Manipulation.}
Table~\ref{tab:additional_tasks_performance} also presents results on \texttt{scene-play}. BGP achieves a mean success rate of 60\%, compared with 38\% for HIQL, 51\% for GCIQL, 5\% for QRL, and 19\% for CRL. This evaluation supplements the navigation experiments with a manipulation task. These results suggest that BGP can also improve goal-reaching performance on this manipulation task.

\begin{table}[ht]
    \centering
    \scriptsize
    \setlength{\tabcolsep}{2pt}
    \renewcommand{\arraystretch}{1.2}
    \providecommand{\best}[1]{\textcolor[rgb]{0.00,0.58,0.62}{\textbf{#1}}}
    \providecommand{\std}[1]{\textsuperscript{($\pm$#1)}}
    \caption{Success rates (\%) on \texttt{humanoidmaze} and \texttt{scene-play-v0}. The highest mean in each row is highlighted in \best{bold}. The final column reports the BGP hyperparameter settings $(\alpha,N)$, with -- indicating an unreported setting.}
    \begin{tabular*}{\textwidth}{@{\extracolsep{\fill}}llcccccc@{}}
        \toprule
        \textbf{Env} & \textbf{Dataset} & \textbf{HIQL} & \textbf{GCIQL} & \textbf{QRL} & \textbf{CRL} & \textbf{BGP(Ours)} & \textbf{BGP} $(\alpha,N)$ \\
        \midrule
        \multirow{6}{*}{\texttt{humanoidmaze}}
        & \texttt{medium-navigate} & 89\std{2} & 27\std{2} & 21\std{8} & 60\std{4} & \best{95\std{2}} & $(10,50)$ \\
        & \texttt{medium-stitch} & \best{88\std{2}} & 12\std{3} & 18\std{2} & 36\std{2} & 87\std{6} & $(10,50)$ \\
        & \texttt{large-navigate} & 49\std{4} & 2\std{1} & 5\std{1} & 24\std{4} & \best{70\std{7}} & $(20,50)$ \\
        & \texttt{large-stitch} & 28\std{3} & 0\std{0} & 3\std{1} & 4\std{1} & \best{58\std{8}} & $(50,50)$ \\
        & \texttt{giant-navigate} & 12\std{4} & 0\std{0} & 1\std{0} & 3\std{2} & \best{75\std{6}} & $(10,150)$ \\
        & \texttt{giant-stitch} & 3\std{2} & 0\std{0} & 0\std{0} & 0\std{0} & \best{69\std{8}} & $(10,150)$ \\
        \midrule
        \texttt{scene} & \texttt{play} & 38\std{3} & 51\std{4} & 5\std{1} & 19\std{2} & \best{60\std{5}} & $(50,150)$ \\
        \bottomrule
    \end{tabular*}
    \label{tab:additional_tasks_performance}
\end{table}

\clearpage
\subsection{Trajectory Analysis}
\label{subsecapp:trajectory_analysis}

To provide additional intuition into the planning behavior of BGP, we visualize representative landmarks and executed trajectories for each layout. Since the qualitative behavior is largely consistent across different dataset variants within the same layout, we present one representative example per layout.

\begin{figure}[h]
    \centering
    \includegraphics[width=0.8\linewidth]{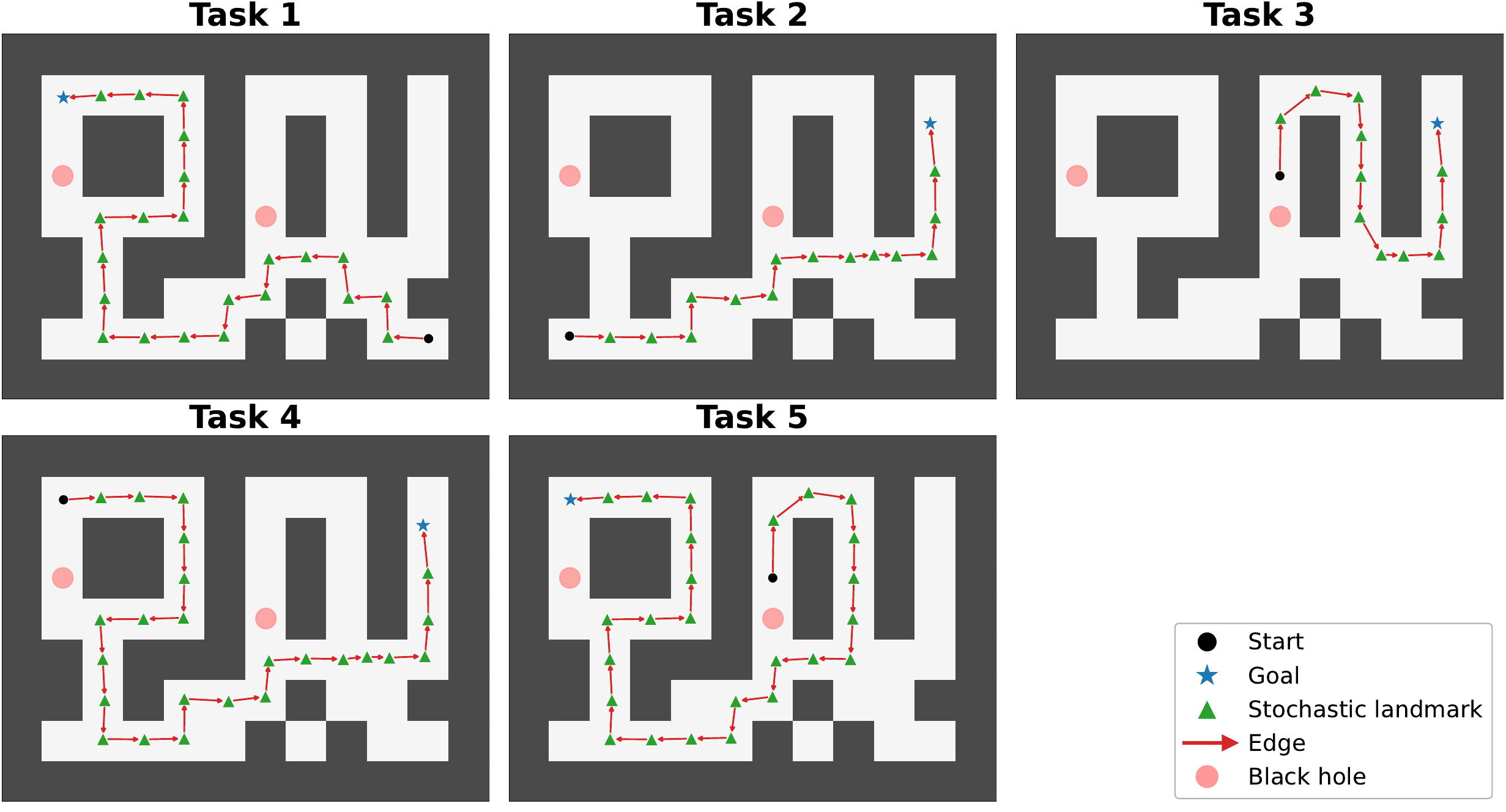}
    \caption{Representative trajectory analysis on the \texttt{teleport} environment.}
    \label{fig:traj_teleport}
\end{figure}

\paragraph{Teleport layout.} First, Fig.~\ref{fig:traj_teleport} illustrates the navigation results in the \texttt{teleport} layout. In these settings, teleportation zones introduce severe and unpredictable state transitions. Methods that rely solely on local connectivity can be misled by these highly branching regions, treating them as spuriously efficient shortcuts. As the visualization demonstrates, BGP correctly identifies these critical branching points and systematically routes the agent around them. By evaluating the expected outcomes of downstream branches, the planner avoids deceptive paths and constructs a reliable route that safely circumvents the teleportation zones.

\begin{figure}[h]
    \centering
    \includegraphics[width=0.8\linewidth]{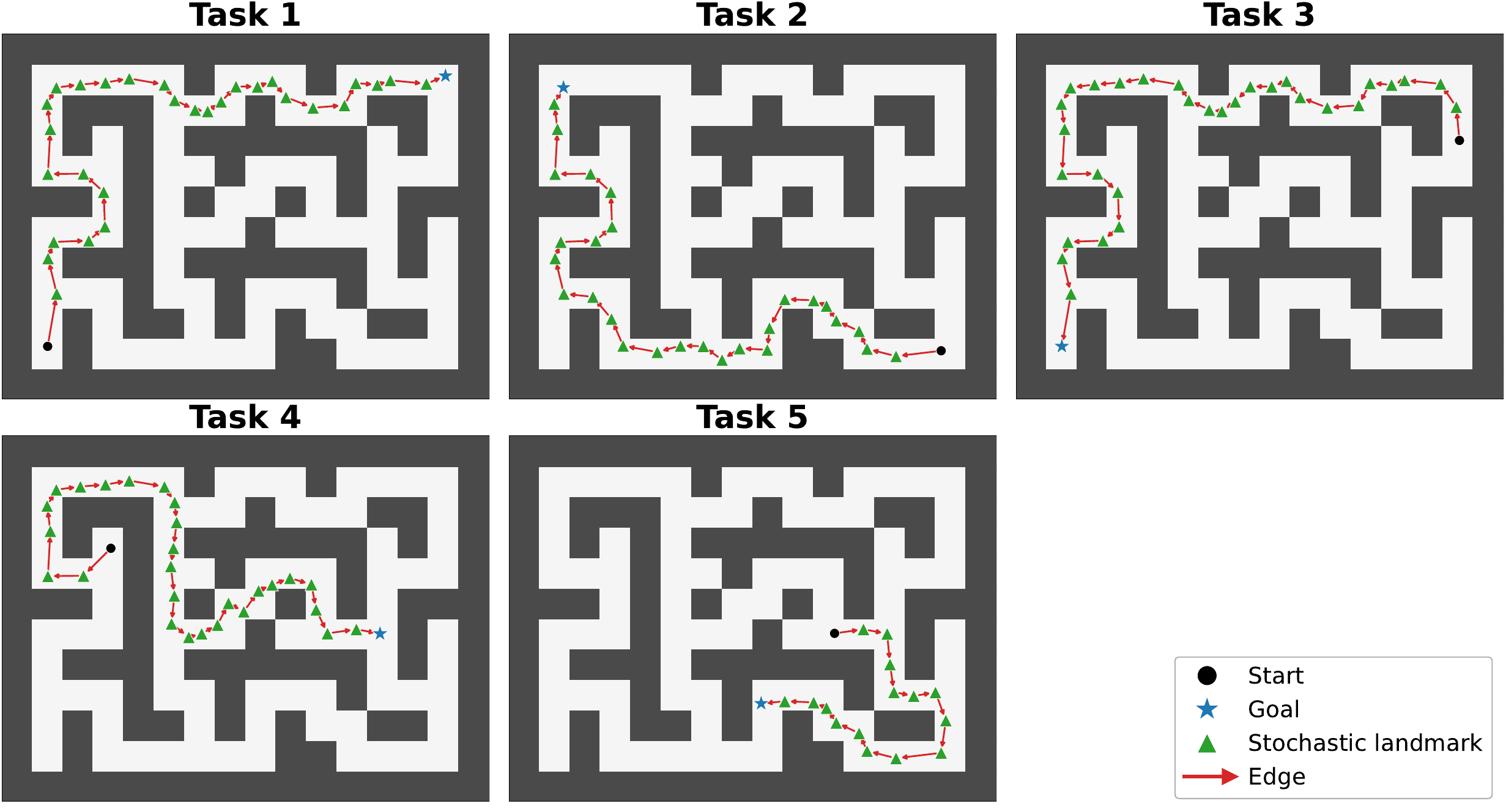}
    \caption{Representative trajectory analysis on the \texttt{giant} environment.}
    \label{fig:traj_giant}
\end{figure}

\paragraph{Giant layout.} Fig.~\ref{fig:traj_giant} shows the corresponding trajectory visualization on the \texttt{giant} layout. This environment contains long maze-like corridors and many branching paths, making long-horizon planning particularly challenging. Despite this complexity, BGP organizes navigation through a sequence of reliable intermediate landmarks, producing a coherent waypoint trajectory toward the distant goal. The resulting graph path demonstrates that BGP effectively decomposes difficult long-range planning problems into stable local transitions.

\begin{figure}[h]
    \centering
    \includegraphics[width=0.8\linewidth]{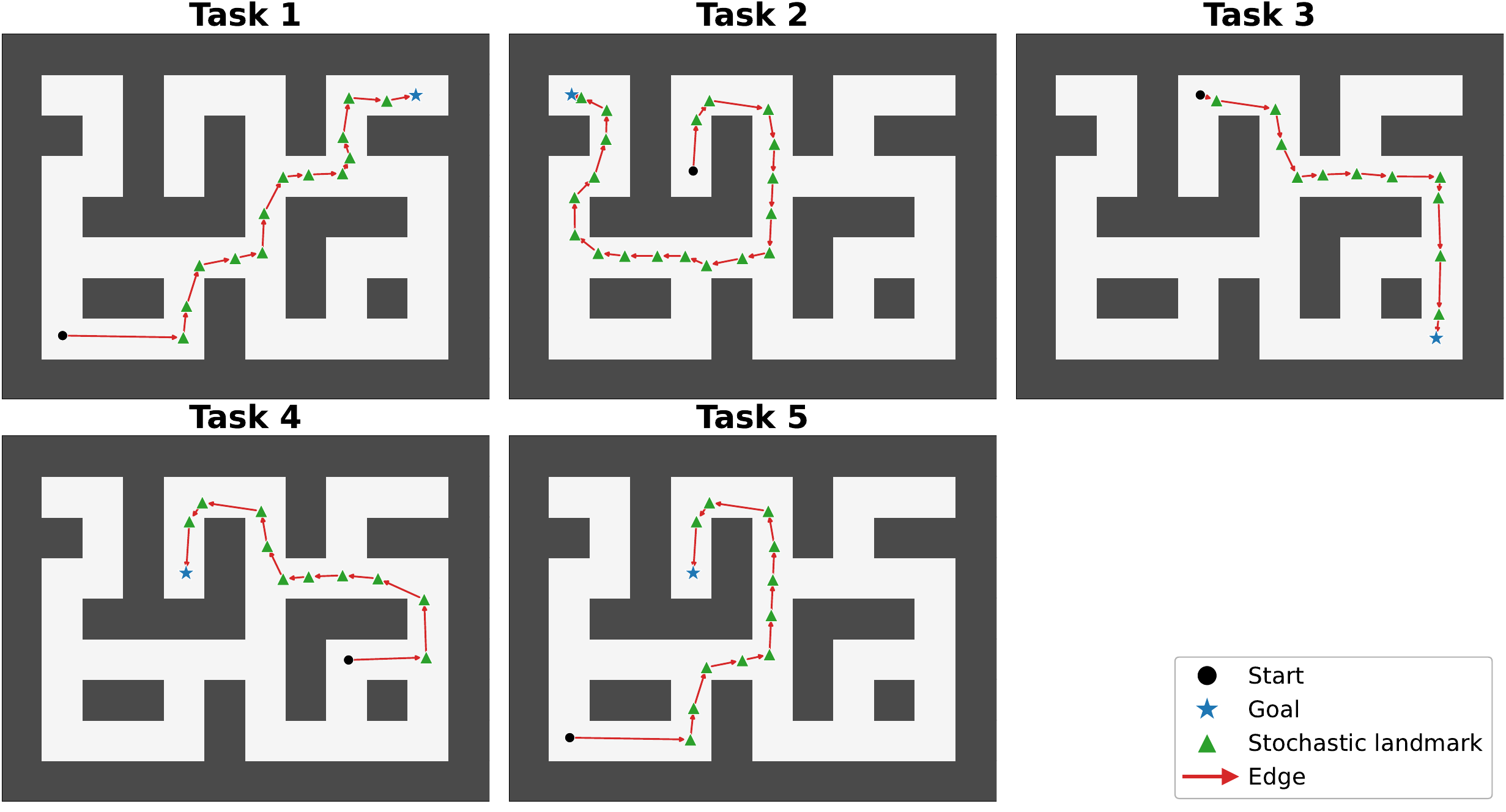}
    \caption{Representative trajectory analysis on the \texttt{large} environment.}
    \label{fig:traj_large}
\end{figure}

\clearpage
\paragraph{Large layout.} Finally, Fig.~\ref{fig:traj_large} presents the \texttt{large} layout. Although this environment is smaller and structurally simpler than the \texttt{giant} layout, it still contains multiple branching regions that require reliable intermediate decisions. BGP consistently identifies these meaningful branching junctions as stochastic landmarks and composes a coherent route toward the target goal. The resulting trajectory demonstrates that the learned graph structure remains informative even in relatively smaller environments, allowing the agent to reliably reach the goal through branch-aware waypoint planning.

\clearpage

\subsection{Hyperparameter Sensitivity}
\label{subsecapp:hyperparameter_sensitivity}

We further analyze the sensitivity of BGP with respect to its key hyperparameters: the percentile threshold $\alpha$, the number of clusters $N$, and the segment length $M$. Overall, while BGP exhibits distinct trends depending on the environmental characteristics, it maintains a generally robust performance across a broad range of configurations.

\begin{figure}[ht]
    \centering
    \includegraphics[width=0.9\linewidth]{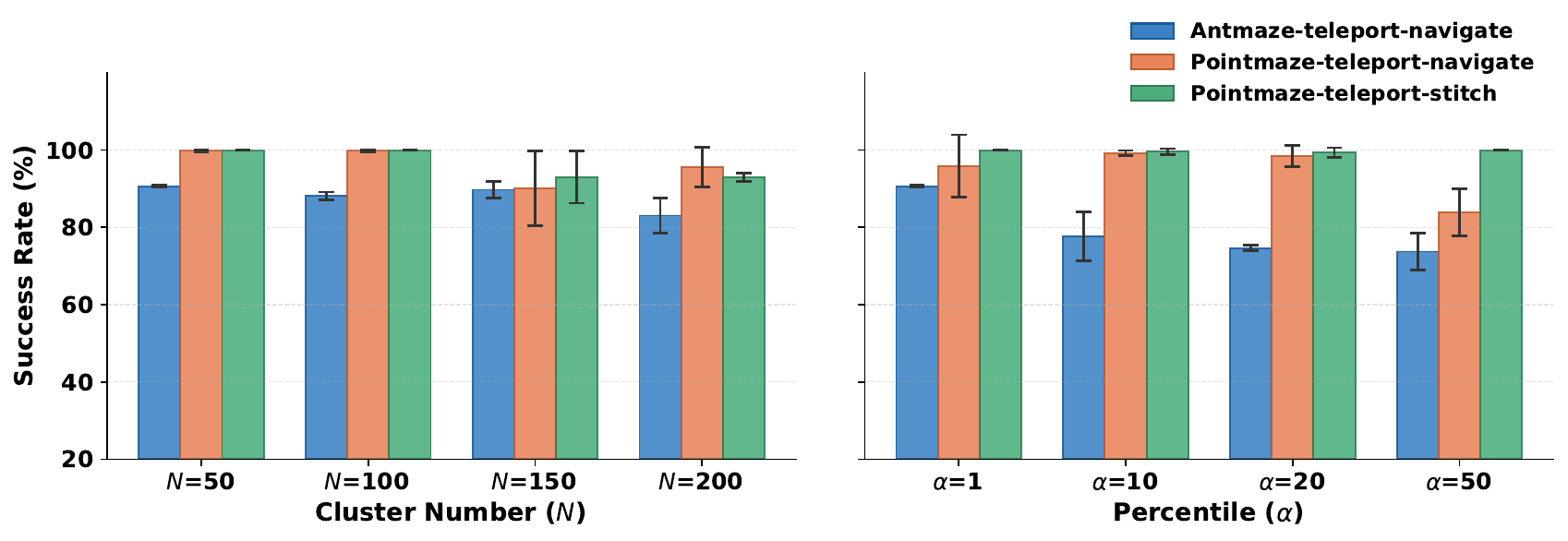}
    \caption{Hyperparameter sensitivity on the \texttt{teleport} environment.}
    \label{fig:teleport_hyperparameter}
\end{figure}

\begin{figure}[ht]
    \centering
    \includegraphics[width=0.9\linewidth]{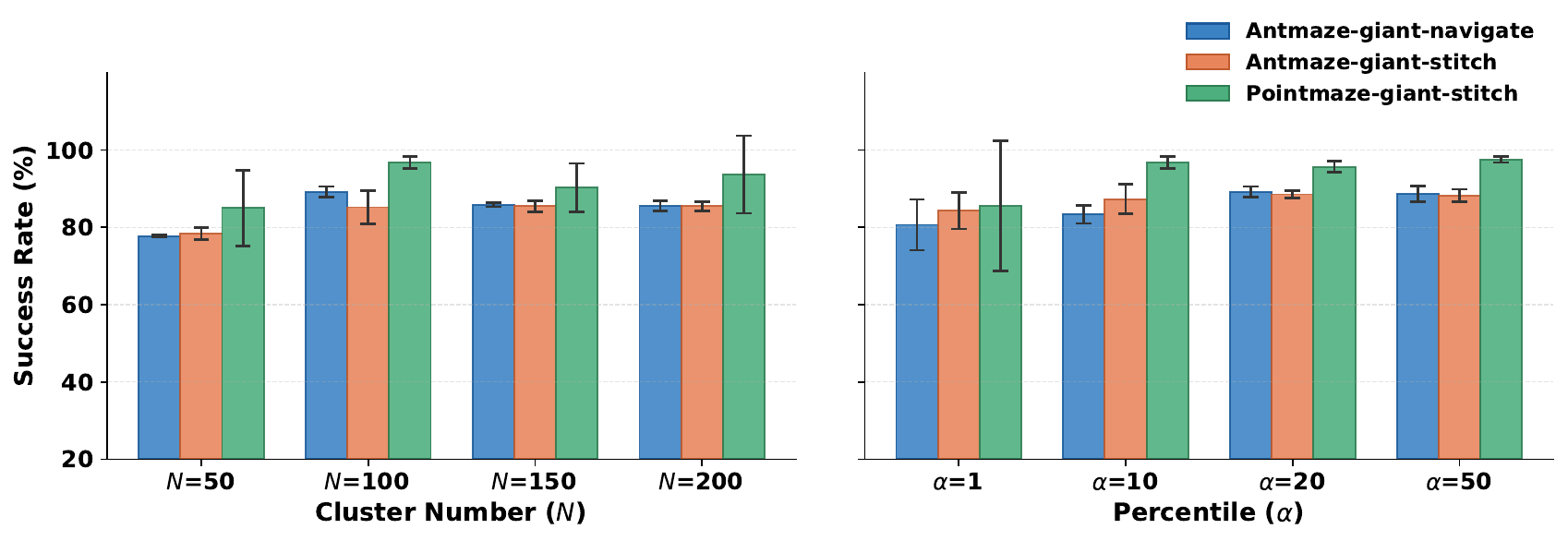}
    \caption{Hyperparameter sensitivity on the \texttt{giant} environment.}
    \label{fig:giant_hyperparameter}
\end{figure}

\paragraph{Landmark parameters ($\alpha$ and $N$).}
Fig.~\ref{fig:teleport_hyperparameter} and Fig.~\ref{fig:giant_hyperparameter} illustrate the success rates across $\alpha$ and $N$ for the \texttt{teleport} and \texttt{giant} environments, respectively. We tested these parameters across multiple environments and observed similar overall trends. Variations in these parameters yield no significant performance differences in pointMaze domains. The performance differences become more pronounced in antMaze domains and are especially clear within the \texttt{teleport} environments. The \texttt{giant} environments also exhibit trends consistent with this overall pattern. Despite these layout-specific tendencies, BGP achieves consistently high success rates across various $(\alpha, N)$ pairs, indicating that it can be reliably deployed without exhaustive tuning.

\paragraph{Segment length ($M$).}

Table~\ref{tab:segment_length_sensitivity} summarizes the performance variations when adjusting the number of segments $M$ within a local training chunk. We evaluate this across both HIQL and GCIQL backbones. If $M$ is too small, the local training data lacks a sufficient horizon to effectively propagate value information between distinct stochastic branches. This prevents the policy from properly learning how to navigate toward waypoints. Conversely, setting the segment length excessively large (e.g., $M=10$) incorporates too many stochastic boundaries within a single training chunk, leading to an accumulation of value estimation bias. As shown in the table, GCIQL exhibits a noticeable performance decline at $M=10$ because its single flat policy is highly sensitive to compounded stochasticity over extended local horizons. HIQL demonstrates greater robustness due to its hierarchical structure but still suffers performance degradation at a larger segment length. We therefore adopt $M=3$ as a default setting that provides enough horizon for value propagation while avoiding the accumulation of estimation bias.

\begin{table}[ht]
    \centering
    \small
    \caption{Sensitivity to the segment length ($M$). Results are averaged over five random seeds. The $M=10$ case excludes \texttt{stitch} datasets, as their inherent trajectory lengths are too short to support this segment length.}
    \label{tab:segment_length_sensitivity}
    \begin{tabular}{lcccc}
        \toprule
        \textbf{Backbone} & $M=2$ & $M=3$ & $M=5$ & $M=10$ (excl. stitch) \\
        \midrule
        HIQL  & $90.0 \pm 2.0$ & $\mathbf{90.2 \pm 1.1}$ & $89.0 \pm 3.2$ & $87.1 \pm 4.0$ \\
        GCIQL & $73.5 \pm 15.6$ & $79.6 \pm 11.4$ & $72.9 \pm 16.4$ & $48.5 \pm 21.7$ \\
        \bottomrule
    \end{tabular}
\end{table}

\newpage

\subsection{Hyperparameter Search Budget}
\label{subsecapp:budget_aware}

To characterize the practical tuning cost of BGP, we adopt the budget-aware evaluation protocol of \citet{jackson2026clean}, which measures performance as a function of the hyperparameter search budget rather than reporting only the best configuration found after an unconstrained search. We evaluate BGP on \texttt{antmaze-teleport-navigate} using a grid of 60 configurations spanning its three hyperparameters, $N \in \{50,100,150,200\}$, $\alpha \in \{0.5,1,10,20,50\}$, and $M \in \{2,3,5\}$, training and evaluating 16 sampled policies for each configuration. We then simulate sequential hyperparameter search using a UCB bandit following \citet{jackson2026clean} and report the expected success rate after a given number of evaluation episodes, averaged over repeated searches. The corresponding 95\% confidence intervals reflect the stochasticity of the bandit's configuration selection.

Table~\ref{tab:budget_aware} shows that BGP achieves an expected success rate of 65\% with a single evaluation, exceeding the per-task tuned CRL (53\%), HIQL (42\%), and GAS (39\%) baselines reported in Table~\ref{tab:ogbench_performance}. Performance increases steadily with the search budget, reaching 80\% after 35 evaluations, which corresponds to approximately 95\% of the asymptotic performance of 84\% obtained after 200 evaluations.

These results show that BGP achieves strong performance under a modest hyperparameter search budget: a single, arbitrarily chosen configuration already outperforms carefully tuned baselines, and the diminishing gains beyond 35 evaluations indicate that extensive search is not necessary to approach its best achievable performance.

\begin{table}[h]
    \centering
    \small
    \caption{Budget-aware evaluation on \texttt{antmaze-teleport-navigate}. We report the expected success rate (\%) and 95\% confidence interval (CI) as a function of the number of evaluation episodes, following \citet{jackson2026clean}.}
    \label{tab:budget_aware}
    \begin{tabular}{lcccccc}
        \toprule
        \textbf{Evaluation episodes} & 1 & 5 & 13 & 35 & 100 & 200 \\
        \midrule
        Success rate (\%) & 65 & 71 & 76 & 80 & 83 & 84 \\
        95\% CI & [12, 88] & [30, 88] & [50, 88] & [62, 88] & [63, 88] & [67, 88] \\
        \bottomrule
    \end{tabular}
\end{table}

\subsection{Computational Cost Analysis}
\label{subsecapp:compute_cost}

\paragraph{Theoretical complexity.}
We provide a theoretical complexity analysis to detail the overhead introduced by BGP. Because the preprocessing steps are executed entirely prior to evaluation, the inference overhead remains highly manageable. Specifically, computing the multi-step predictive uncertainty score requires $\mathcal{O}(|\mathcal{D}|)$ forward passes over the dataset $\mathcal{D}$. Extracting $N$ landmarks from the high-uncertainty states $U$ via K-means++ clustering incurs $\mathcal{O}(U \cdot N \cdot I)$ time complexity, where $I$ is the number of iterations. Constructing the branch-aware edges takes $\mathcal{O}(N^2 \cdot C)$, where $C$ is the bounded number of operations required to aggregate the values of reachable clusters. During evaluation, finding the optimal waypoint path via Dijkstra's algorithm requires $\mathcal{O}(N \log N + |\mathcal{E}|)$ operations, where $|\mathcal{E}|$ is the number of edges. Because the number of landmarks $N$ is deliberately kept small ($50 \le N \le 100$), this theoretical inference overhead is practically negligible and enables fast waypoint generation.

\paragraph{Empirical simulation time.}
To complement the theoretical analysis, Table~\ref{tab:computational_cost} reports the total empirical simulation time across different methods. As expected, implicit value-learning baselines such as GCIQL, HIQL, and CGCIVL exhibit lower computational costs because they do not construct explicit graphical structures or perform path searches. BGP introduces an expected overhead due to uncertainty estimation, clustering, and Dijkstra's algorithm, resulting in a moderate increase in total simulation time. However, BGP is significantly more computationally efficient than graph-based planning methods such as GAS, while also maintaining competitive efficiency relative to temporal-distance-based approaches such as TMD. This demonstrates that BGP achieves a favorable balance, providing robust long-horizon planning in stochastic environments without suffering from the computational overhead often observed in explicit graph-based planning methods.

\begin{table}[h]
    \centering
    \small
    \caption{Simulation time across different methods.}
    \label{tab:computational_cost}
    \begin{tabular}{lcccccc}
        \toprule
        \textbf{Method} & BGP & HIQL & GCIQL & CGCIVL & TMD & GAS  \\
        \midrule
        \textbf{Time (s)} & 7563 & 4686 & 3632 & 4927 & 17577 & 8841 \\
        \bottomrule
    \end{tabular}
\end{table}

\section{Broader Impact}
\label{secapp:broader_impact}
The BGP framework enhances offline goal-conditioned reinforcement learning for complex real-world applications, such as autonomous logistics, by improving planning reliability in stochastic domains. Its explicit graph construction also improves interpretability by enabling human operators to verify routes prior to execution. However, BGP heavily depends on offline dataset quality, and its preprocessing steps introduce computational overhead. Beyond these technical limitations, the framework does not pose unique societal or ethical risks distinct from standard offline reinforcement learning and requires only standard responsible oversight.

\newpage

\section*{NeurIPS Paper Checklist}

\begin{enumerate}

\item {\bf Claims}
    \item[] Question: Do the main claims made in the abstract and introduction accurately reflect the paper's contributions and scope?
    \item[] Answer: \answerYes{} 
    \item[] Justification: We clearly state the contributions and scope of the proposed algorithm, BGP, in the abstract and introduction.

\item {\bf Limitations}
    \item[] Question: Does the paper discuss the limitations of the work performed by the authors?
    \item[] Answer: \answerYes{} 
    \item[] Justification: We discuss the limitations of BGP, especially regarding additional hyperparameters and computational cost, in Section~\ref{sec:limitation}.

\item {\bf Theory assumptions and proofs}
    \item[] Question: For each theoretical result, does the paper provide the full set of assumptions and a complete (and correct) proof?
    \item[] Answer: \answerNA{} 
    \item[] Justification: This paper does not include formal theoretical claims that require mathematical proofs.

    \item {\bf Experimental result reproducibility}
    \item[] Question: Does the paper fully disclose all the information needed to reproduce the main experimental results of the paper to the extent that it affects the main claims and/or conclusions of the paper (regardless of whether the code and data are provided or not)?
    \item[] Answer: \answerYes{} 
    \item[] Justification: We describe the detailed methodologies in Section~\ref{sec:method} and provide further implementation details in Appendix~\ref{secapp:implementation_details}. These descriptions are sufficient to reproduce the main experimental results.

\item {\bf Open access to data and code}
    \item[] Question: Does the paper provide open access to the data and code, with sufficient instructions to faithfully reproduce the main experimental results, as described in supplemental material?
    \item[] Answer: \answerYes{} 
    \item[] Justification: We provide the full source code and running scripts, along with detailed instructions for reproducing all main results, in the supplemental material.

\item {\bf Experimental setting/details}
    \item[] Question: Does the paper specify all the training and test details (e.g., data splits, hyperparameters, how they were chosen, type of optimizer) necessary to understand the results?
    \item[] Answer: \answerYes{} 
    \item[] Justification: We describe the experimental setup, including hyperparameters, datasets, optimizers, etc., in Section~\ref{sec:experimental setup} and Appendix~\ref{secapp:implementation_details}.

\item {\bf Experiment statistical significance}
    \item[] Question: Does the paper report error bars suitably and correctly defined or other appropriate information about the statistical significance of the experiments?
    \item[] Answer: \answerYes{} 
    \item[] Justification:  The main benchmark results report mean $\pm$ standard deviation over five seeds; the stochasticity and budget-aware analyses report 95$\%$ confidence intervals, as specified in the paper.

\item {\bf Experiments compute resources}
    \item[] Question: For each experiment, does the paper provide sufficient information on the computer resources (type of compute workers, memory, time of execution) needed to reproduce the experiments?
    \item[] Answer: \answerYes{} 
    \item[] Justification: We describe the detailed computational setup, including GPU types, memory, and operating system, in Appendix~\ref{subsecapp:environment_detail}. 
    
\item {\bf Code of ethics}
    \item[] Question: Does the research conducted in the paper conform, in every respect, with the NeurIPS Code of Ethics \url{https://neurips.cc/public/EthicsGuidelines}?
    \item[] Answer: \answerYes{} 
    \item[] Justification: The research is based on a standard benchmark, OGBench~\citep{park2024ogbench}, which complies with the NeurIPS Code of Ethics.

\item {\bf Broader impacts}
    \item[] Question: Does the paper discuss both potential positive societal impacts and negative societal impacts of the work performed?
    \item[] Answer: \answerYes{} 
    \item[] Justification: We discuss broader impacts in Appendix~\ref{secapp:broader_impact}.
    
\item {\bf Safeguards}
    \item[] Question: Does the paper describe safeguards that have been put in place for responsible release of data or models that have a high risk for misuse (e.g., pre-trained language models, image generators, or scraped datasets)?
    \item[] Answer: \answerNA{} 
    \item[] Justification: This research does not pose a high risk of misuse or dual-use concerns.

\item {\bf Licenses for existing assets}
    \item[] Question: Are the creators or original owners of assets (e.g., code, data, models), used in the paper, properly credited and are the license and terms of use explicitly mentioned and properly respected?
    \item[] Answer: \answerYes{} 
    \item[] Justification: All environments and baseline implementations were used under their official licenses and are properly cited in this paper.

\item {\bf New assets}
    \item[] Question: Are new assets introduced in the paper well documented and is the documentation provided alongside the assets?
    \item[] Answer: \answerYes{} 
    \item[] Justification: All introduced methods, including algorithms and models, are described in the paper and the provided supplementary material.

\item {\bf Crowdsourcing and research with human subjects}
    \item[] Question: For crowdsourcing experiments and research with human subjects, does the paper include the full text of instructions given to participants and screenshots, if applicable, as well as details about compensation (if any)? 
    \item[] Answer: \answerNA{} 
    \item[] Justification: This research does not include any human subjects or crowdsourcing components.

\item {\bf Institutional review board (IRB) approvals or equivalent for research with human subjects}
    \item[] Question: Does the paper describe potential risks incurred by study participants, whether such risks were disclosed to the subjects, and whether Institutional Review Board (IRB) approvals (or an equivalent approval/review based on the requirements of your country or institution) were obtained?
    \item[] Answer: \answerNA{} 
    \item[] Justification: This research does not involve human subjects or require IRB approval.

\item {\bf Declaration of LLM usage}
    \item[] Question: Does the paper describe the usage of LLMs if it is an important, original, or non-standard component of the core methods in this research? Note that if the LLM is used only for writing, editing, or formatting purposes and does \emph{not} impact the core methodology, scientific rigor, or originality of the research, declaration is not required.
    \item[] Answer: \answerNA{} 
    \item[] Justification: LLMs were used strictly for writing and editing purposes and were not used in the development of the core methodology.

\end{enumerate}

\end{document}